\documentclass[fleqn,10pt]{wlscirep}
\usepackage[utf8]{inputenc}
\usepackage[T1]{fontenc}
\usepackage{comment}

\title{Product Progression: a machine learning approach to forecasting industrial upgrading}

\author[1,2]{Giambattista Albora}
\author[2]{Luciano Pietronero}
\author[3]{Andrea Tacchella}
\author[4,*]{Andrea Zaccaria}
\affil[1]{Dipartimento di Fisica, Università Sapienza, Rome, Italy}
\affil[2]{Centro Ricerche Enrico Fermi, Rome, Italy}
\affil[3]{Joint Research Centre, Seville, Spain}
\affil[4]{Istituto dei Sistemi Complessi - CNR, UOS Sapienza, Rome, Italy}

\affil[*]{andrea.zaccaria@cnr.it}

\begin{abstract}
Economic complexity methods, and in particular relatedness measures, lack a systematic evaluation and comparison framework. We argue that out-of-sample forecast exercises should play this role, and we compare various machine learning models to set the prediction benchmark. We find that the key object to forecast is the activation of new products, and that tree-based algorithms clearly overperform both the quite strong auto-correlation benchmark and the other supervised algorithms. Interestingly, we find that the best results are obtained in a cross-validation setting, when data about the predicted country was excluded from the training set. Our approach has direct policy implications, providing a quantitative and scientifically tested measure of the feasibility of introducing a new product in a given country.
\end{abstract}
\begin{document}

\flushbottom
\maketitle
%
%
    \thispagestyle{empty}

\section*{Introduction}

In her essay \textit{The Impact of Machine Learning on Economics}, Susan Athey states: ``Prediction tasks [...] are typically not the problems of greatest interest for empirical research in economics, who instead are concerned with causal inference '' and 
``economists typically abandon the goal of accurate prediction of outcomes in pursuit of an unbiased estimate of a causal parameter of interest '' \cite{athey2018impact}.  This situation is mainly due to two factors: the need to ground policy prescriptions \cite{rodrik2010diagnostics,hausmann2008growth} and the intrinsic difficulty to make correct predictions in complex systems \cite{baldovin2018role,hosni2018forecasting}. The immediate consequence of this behaviour is the flourishing of different or even contrasting economic models, whose concrete application largely relies on the specific skills, or biases, of the scholar or the policymaker \cite{rodrik2015economics}. This \textit{horizontal} view, in which models are every time aligned and selected, in contrast with the \textit{vertical} view of hard sciences, in which models are selected by comparing them with empirical evidence, leads to the challenging issue of distinguishing which models are wrong. While this situation can be viewed as a natural feature of economic and, more in general, complex systems \cite{rodrik2015economics}, a number of scholars coming from hard sciences have recently tackled these issues, trying to introduce concepts and methods from their disciplines in which models' falsifiability, tested against empirical evidence, is \textit{the} key element. This innovative approach, called \textit{Economic Fitness and Complexity}\cite{Tacchella,cristelli2013measuring,tacchella2013economic,tacchella2018dynamical,zaccaria2014taxonomy,zaccaria2016case} (EFC), combines statistical physics and complex network based algorithms to investigate macroeconomics with the aim to provide testable and scientifically valid results. The EFC methodology studies essentially two lines of research: indices for the competitiveness of countries and relatedness measures. \\The first one aims at assessing the industrial competitiveness of countries by applying iterative algorithms to the bipartite network connecting countries to the products they competitively export \cite{gaulier2010baci}. Two examples are the Economic Complexity Index ECI \cite{hidalgo2009building} and the Fitness \cite{Tacchella}. In this case the scientific soundness of either approach can be assessed by accumulating pieces of evidence: by analyzing the mathematical formulation of the algorithm and the plausibility of the resulting rankings \cite{albeaik1707improving,gabrielli2017we,albeaik1708729,pietronero2017economic}, and by using the indicator to predict other quantities. In particular, the Fitness index, when used in the so-called Selective Predictability Scheme \cite{cristelli2015heterogeneous}, provides GDP growth predictions that over-perform the ones provided by the International Monetary Fund \cite{cristelli2017predictability,tacchella2018dynamical}. All these elements concur towards the plausibility of the Fitness approach; however, a \textit{direct} way to test the predictive performance of these indicators\cite{liao2018comparative} is still lacking. This naturally leads to the consideration of further indices, that can mix the existing ones \cite{sciarra2020reconciling} or use new concepts such as information theory \cite{frenken2007related}. We argue that, on the contrary, the scientific validity of relatedness indicators can be assessed, and this is the purpose of the present work.\\
The second line of research in EFC investigates the concept of Relatedness \cite{hidalgo2018principle}, the idea that two human activities are, in a sense, \textit{similar} if they share many of the capabilities needed to be competitive in them \cite{teece1994understanding}. Practical applications are widespread and include international trade \cite{hidalgo2007product,zaccaria2014taxonomy}, firm technological diversification \cite{breschi2003knowledge,pugliese2019coherent}, regional smart specialization \cite{neffke2011regions,boschma2012technological}, and the interplay among the scientific, technological, and industrial layers \cite{pugliese2019unfolding}. Most of these contributions use relatedness not to forecast future quantities, but as an independent variable in a regression, and so the proximity (or quantities derived from it) is used to explain some observed simultaneous behavior. We point out, moreover, that no shared definition of relatedness exists, except for the widespread use of co-occurrences, since different scholars use different normalizations, null models, and data, so the problem to decide ``which model is wrong'' persists.\\
Given this context, we believe that it is fundamental to introduce elements of falsifiability in order to compare the different existing models, and that such comparison should be made by looking at the performance in out-of-sample forecasting, that is the focus of the present paper. We will consider export as the economic quantity to forecast because most of the indicators used in economic complexity are derived from export data, being it regarded as a global, summarizing quantity of countries' capabilities \cite{hausmann2007you,tacchella2018dynamical} but also for the immediate policy implications of the capability to be able, for instance, to predict in which industrial sector a country will be competitive, say, in five years. \\
In this paper we propose a procedure to systematically compare different prediction approaches and, as a consequence, to scientifically validate or falsify the underlying models. Indeed, some attempts to use complex networks or econometric approaches to predict exports exist \cite{o2021productive,bustos2012dynamics,medo2018link,zhang2019industry}, but these methodologies are practically impossible to compare precisely because of the lack of a common framework to choose how to preprocess data, how to build the training and the test set, or even which indicator to use to evaluate the predictive performance. In the following, we will systematically scrutiny the steps to build a scientifically sound testing procedure to predict the evolution of the export basket of countries. In particular, we will forecast the presence or the activation of a binary matrix element $M_{cp}$, that indicates whether the country $c$ competitively exports product $p$ in the Revealed Comparative Advantage sense \cite{balassa1965trade} (see Methods for a detailed description of the export data).  Given the present ubiquitous and successful use of artificial intelligence in many different contexts, it is natural to use machine learning algorithms to set the benchmark. A relevant by-product of this analysis is the investigation of the relationships between the statistical properties of the database (namely, a strong auto-correlation and class imbalance) and the choice of the most suitable algorithms and performance indicators.\\
Applying these methods we find two interesting results:
\begin{enumerate}
    \item The best performing models for this task are based on decision trees. A fundamental property that separates these algorithms from the main approaches used in the literature \cite{hidalgo2007product} is the fact that here the presence of a product in the export basket of a country can have a negative effect in the probability of exporting the target product. I.e. decision trees are able to combine Relatedness and Anti-Relatedness signals to provide strong improvements in the accuracy of predictions\cite{tacchella2021relatedness}
    \item Contrary to what is commonly observed with machine learning models, our best model performs better in a cross-validation setting where we exclude data from the predicted country from the training set. We interpret this surprising finding by arguing that in cross-validation the model is able to better learn the actual Relatedness relationships among products, rather than focusing on the very strong self-correlation of the trade data.
\end{enumerate}
The present investigation of the predictability of the time evolution of export baskets has a number of practical and theoretical applications. First, predicting time evolution of the export basket of a country needs, as an intermediate step, an assessment of the likelihood that the single product will be competitively exported by the country in the next years. This likelihood can be seen as a measure of the \textit{feasibility} of that product, given the present situation of that country. The possibility to investigate with such a great level of detail which product is relatively \textit{close} to a country and which one is out of reach has immediate implications in terms of strategic policies \cite{hausmann2019roadmap}. Secondly, the study of the time evolution of the country-product bipartite network is key to validate the various attempts to model it \cite{saracco2015innovation,tacchella2016build}. Finally, the present study represents one of the first attempts to systematically investigate how machine learning techniques can be applied in development economics, creating a bridge between two previously unrelated disciplines.\\

\section*{Results}
\subsection*{Statistical properties of the country-product network}
\begin{figure}[h]
\centering
\includegraphics[width=0.48\linewidth]{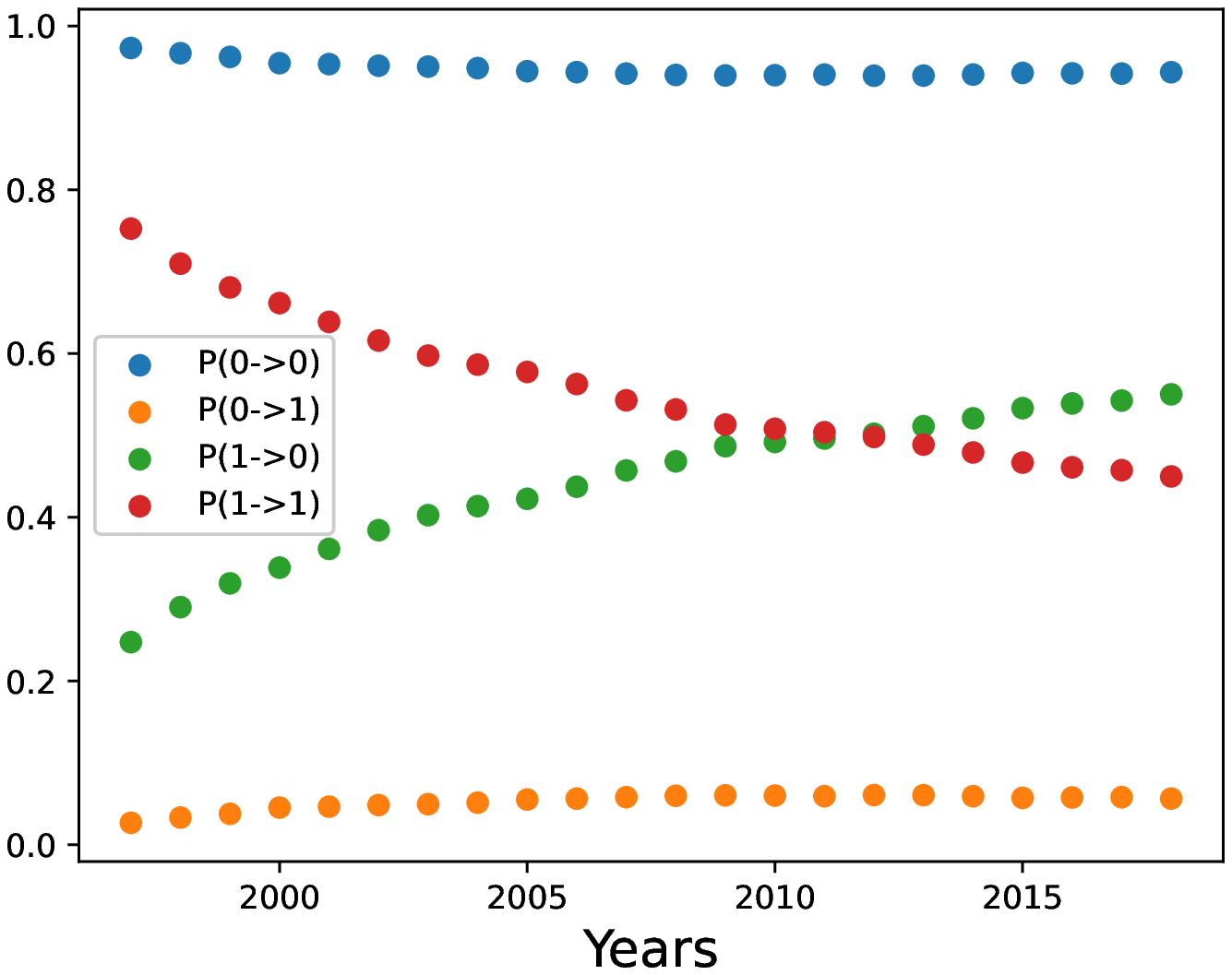}
\includegraphics[width=0.48\linewidth]{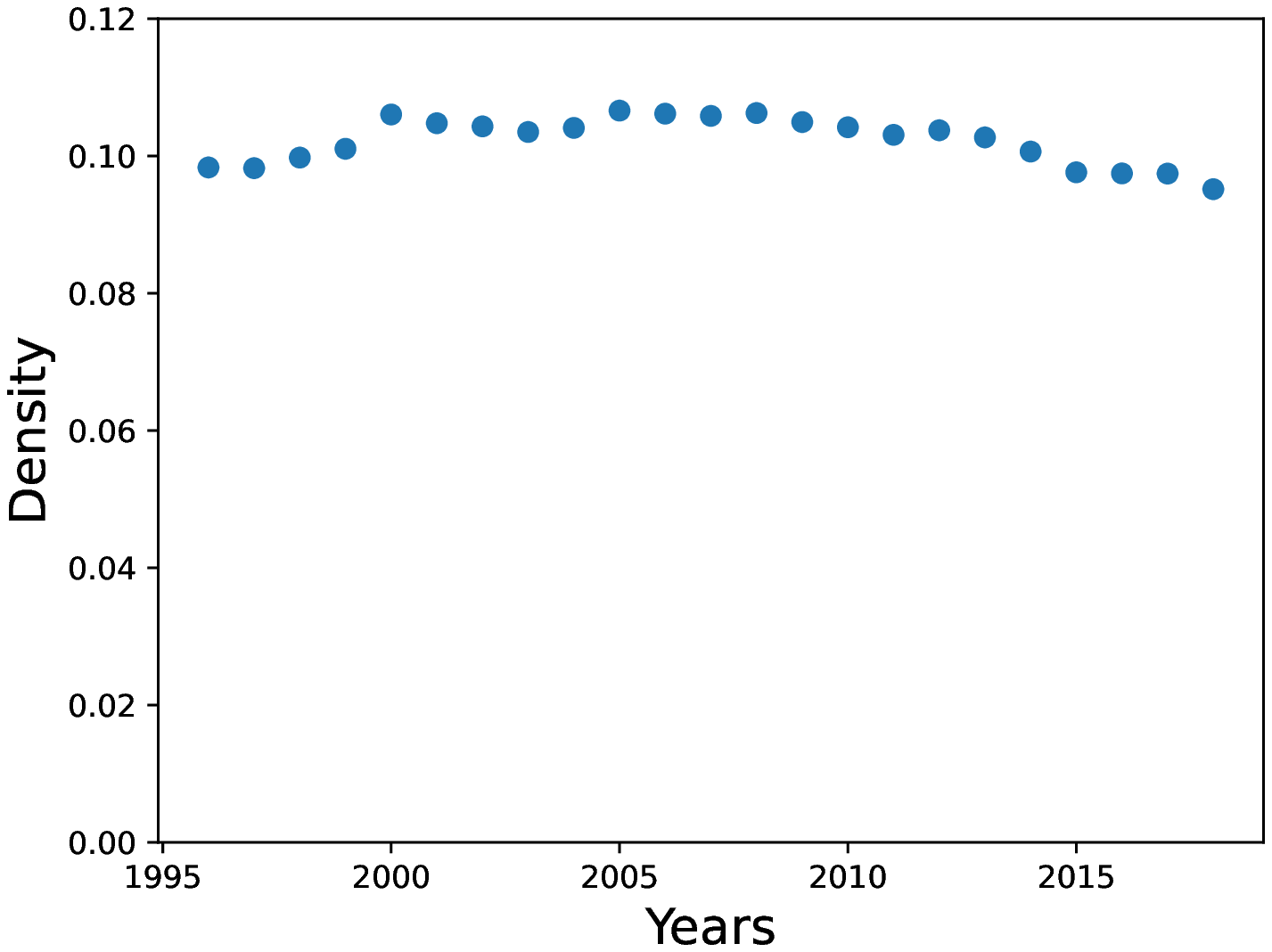}
\caption{Left: transition probabilities between the binary states of the export matrix \textbf{M}. The strong persistency implies the importance of the study of the appearance of new products with respect to the unconditional presence of one matrix element. Right: the fraction of nonzero elements in \textbf{M} as a function of time. A strong class imbalance is present.}
\label{fig:prob}
\end{figure}
A key result of the present investigation is a clear-cut methodology to compare different models or predictive approaches in Economic Complexity. In order to understand the reasons behind some of the choices we made in building the framework, we first discuss some statistical properties of the data we will analyse.\\
Our database is organized in a set of matrices $\textbf{V}$ whose element $V_{cp}$ is the amount, expressed in US dollars, of product $p$ exported by country $c$ in a given year. When not otherwise specified, the number of countries is 169, the number of products is 5040, and the time range covered by our analysis is 1996-2018. We use the HS1992, 6-digits classification. The data are obtained from the UN-COMTRADE database and suitably sanitized in order to take into account the possible disagreements between importers and exporters. We compute the Revealed Comparative Advantage \cite{balassa1965trade} to obtain a set of RCA matrices $\textbf{R}$ and, by applying a threshold equal to 1, a set of matrices $\textbf{M}$ whose binary elements are equal to 1 if the given country competitively exports the given product. Here and in the following we use ``competitively'' in the Balassa sense, that is, $R_{cp}>1$. In this paper we will discuss the prediction of two different situations: the unconditional presence of a ``1'' element in the $\textbf{M}$ matrix and the \textit{appearance} of such an element requiring that the RCA values were below a non-significancy threshold t=0.25 in all the previous years. We will refer to the first case as the \textit{full matrix} and to the new product event as an \textit{activation}. The definition of the activation is somehow arbitrary: one could think, for instance, to change the threshold t or the number of inactive years. We find however our choice to be a good trade-off to have both a good numerosity of the test set and avoid the influence of trivial 0/1 flips. We point out that our final aim is to detect, as much as possible, the appearance of really new products in the export basket of countries.
\\
In Fig.\ref{fig:prob}, left, we plot the probability that a matrix element $M_{cp}$ in 1996 will change or not change its binary value in the future years. One can easily see that even after 5 years the probability that a country remains competitive in a product is relatively high ($\sim 0.64$); being the probability that a country remains not competitive $\sim 0.95$, we conclude that there is a very strong auto-correlation: this is a reflection of the persistent nature of both the capabilities and the market conditions that are needed to competitively export a product. Moreover, the appearance of a new product in the export basket of a country is a rare event: the empirical frequency is about $0.047$ after 5 years. 
A consequence of this persistency is that we can safely predict the presence of a 1 in the $\textbf{M}$ matrices by simply looking at the previous years, while the appearance of a new product that was not previously exported by a country is much more difficult and, in a sense, more interesting from an economical point of view, since it depends more on the presence of suitable, but unrevealed, capabilities in the country; but these capabilities can be traced by looking at the other products that country exports. Not least, an early detection of a future activation of a new product has a number of practical policy implications. Note in passing that, since machine learning based smoothing procedures \cite{tacchella2018dynamical,angelini2018complexity} may introduce extra spurious correlations, they should be avoided in predictions exercises, and so only the RCA values directly computed from the raw export data are considered.
\\
On the right side of Fig.\ref{fig:prob} we plot the density of the matrices $\textbf{M}$, that is the number of nonzero elements with respect to the total number of elements. This ratio is roughly $10\%$.
This means that both the prediction of the \textit{full}, unconditional matrix elements and the prediction of the so-called \textit{activations} (i.e., conditioning to that element being 0 and with RCA below 0.25 in all the previous years) show a strong class imbalance. This has deep consequences regarding the choice of the performance indicators to compare the different predictive algorithms. For instance, the ROC-AUC score \cite{fawcett2006introduction}, one of the most used measure of performance for binary classifiers, is well known to suffer from strong biases when a large class imbalance is present \cite{saito2015precision}. More details are provided in the Methods sections.\\
\subsection*{Recognize the country vs. learning the products' relations}
In this section we present the results concerning the application of different supervised learning algorithms. The training and the test procedures are fully described in the Methods section. Here we just point out that the  training set is composed by the matrices $\textbf{R}^{(y)}$ with $y \in [1996\dots2013]$, and the test is performed against $\textbf{M}^{(2018)}$, so we try to predict the export basket of countries after $\Delta=5$ years.\\
The algorithms we tested are XGBoost \cite{friedman2001greedy,chen2016xgboost}, a basic Neural Network implemented using the Keras library \cite{gulli2017deep} and the following algorithms implemented using the scikit learn library \cite{pedregosa2011scikit}: Random Forest \cite{breiman2001random}, Support Vector Machines \cite{cortes1995support}, Logistic Regression \cite{hosmer2013applied}, a Decision Tree \cite{quinlan1986induction}, Extra Trees Classifiers \cite{geurts2006extremely}, ADA Boost \cite{freund1997decision} and Gaussian Naive Bayes \cite{john2013estimating}. For reasons of space, we cannot discuss all these methods here. However, a detailed description can be found in \cite{shalev2014understanding} and references therein and, in the following sections, we will elaborate more on the algorithms based on decision trees, which result to be the most performing ones.\\
\begin{figure}[ht]
\centering
\includegraphics[width=0.48\linewidth]{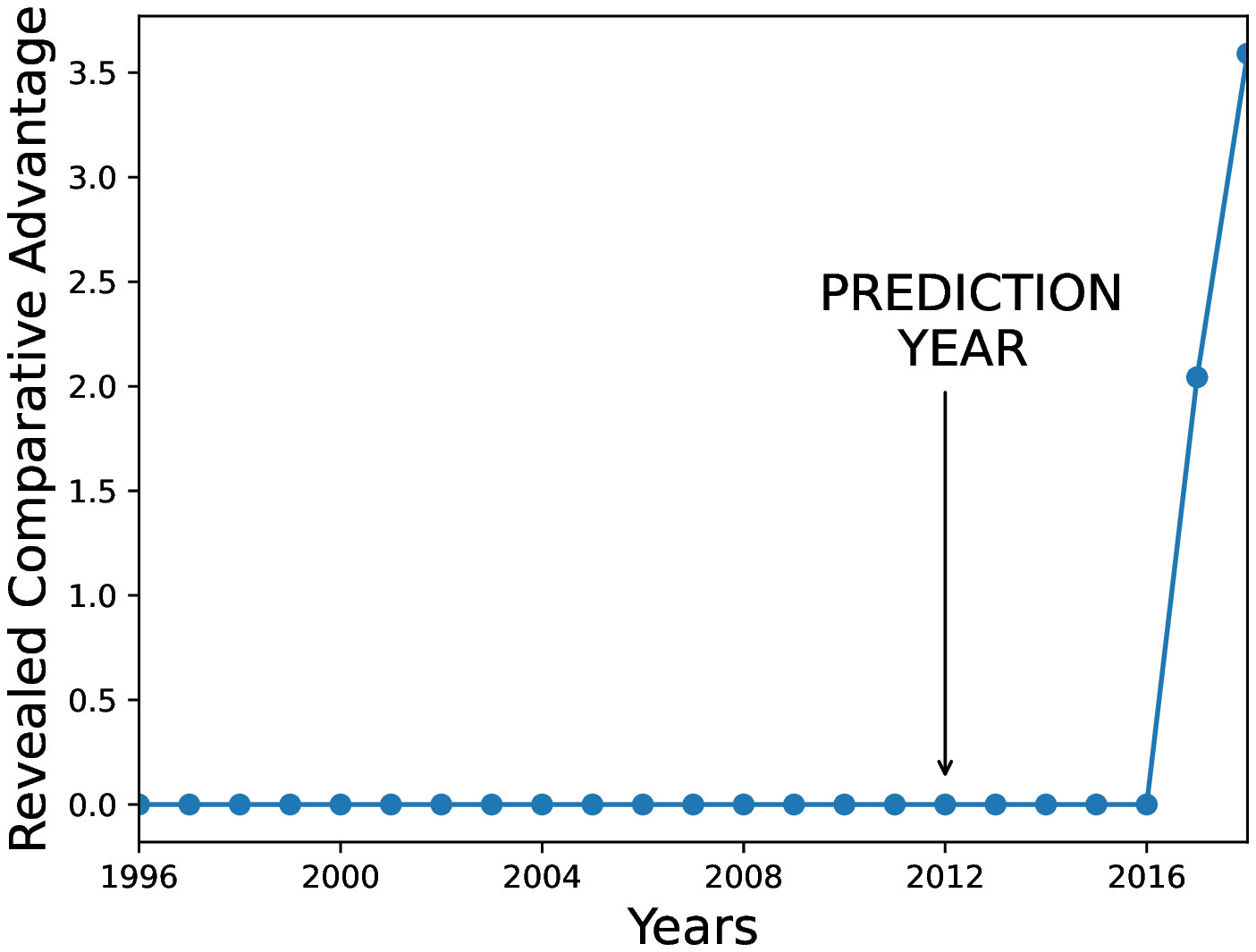}
\includegraphics[width=0.48\linewidth]{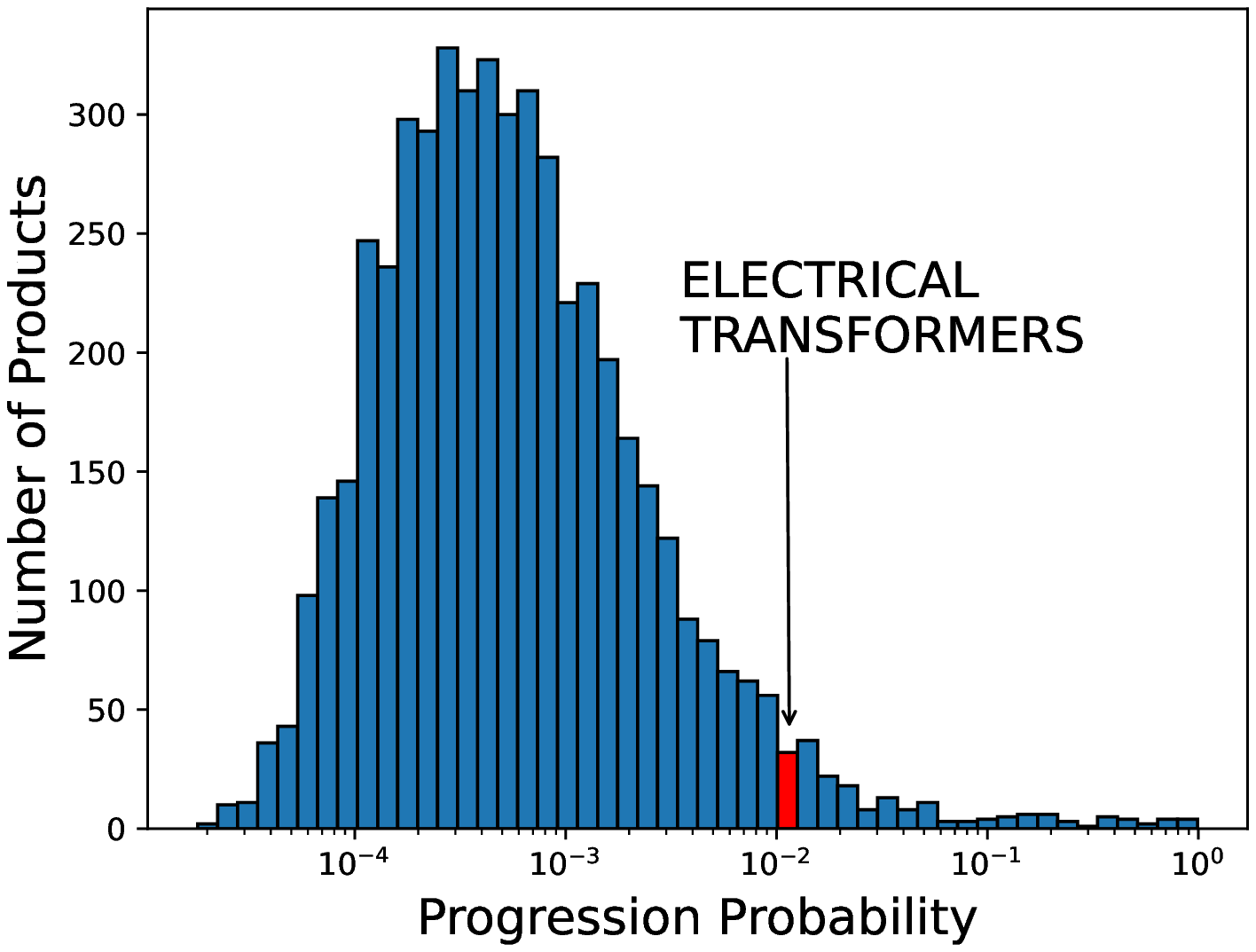}
\caption{An example of successful prediction. On the left, the RCA of Bhutan in electrical transformers as a function of time. Already in 2012, with RCA stably below 1, the progression probability of that matrix element was well above its country average (figure on the right). Bhutan will start to competitively export electrical transformers after 5 years.}
\label{fig:manioca}
\end{figure}
In Fig.\ref{fig:manioca} we show an example of the dynamics that our approach is able to unveil. On the left we show the RCA of Bhutan for the export of Electrical Transformers. RCA is zero from 1996 to 2016, when a sharp increase begins. If we apply the XGBoost algorithm to predict which products will Bhutan likely export in the future, we obtain a set of scores, or \textit{progression probabilities}, one score for each possible product. The distribution of such scores are depicted in Fig.\ref{fig:manioca} on the right. The progression probability for Electrical Transformers was much higher than average, as shown by the arrow. Indeed, as shown by the figure on the left, Bhutan will start to export that specific product in about five years. Obviously, this is just an example, so we need a set of tool to measure the performance on the whole test set on a statistical basis.\\ 
\begin{figure}[ht!]
\centering
\includegraphics[width=0.42\linewidth]{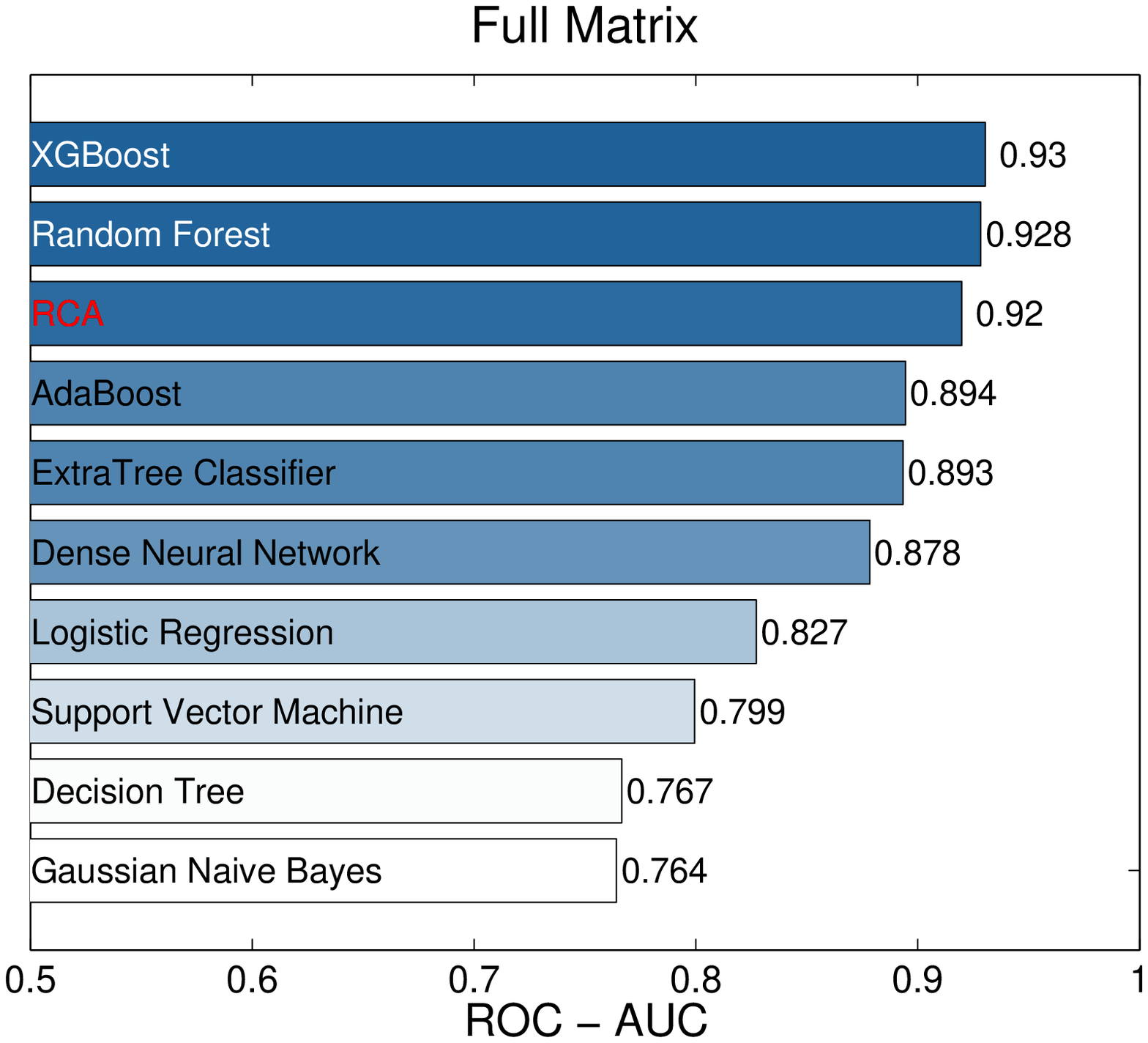}
\includegraphics[width=0.42\linewidth]{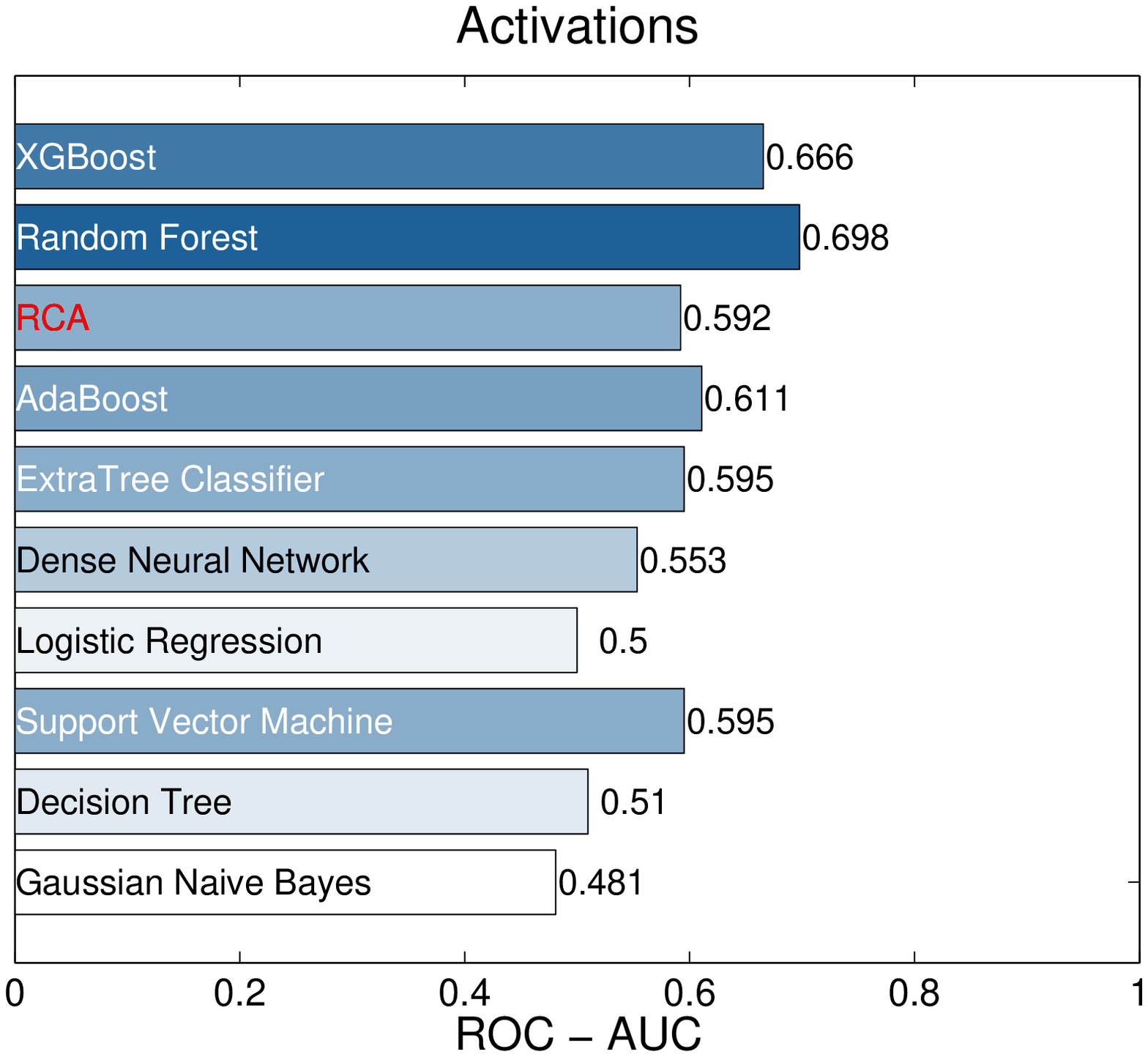}
\includegraphics[width=0.42\linewidth]{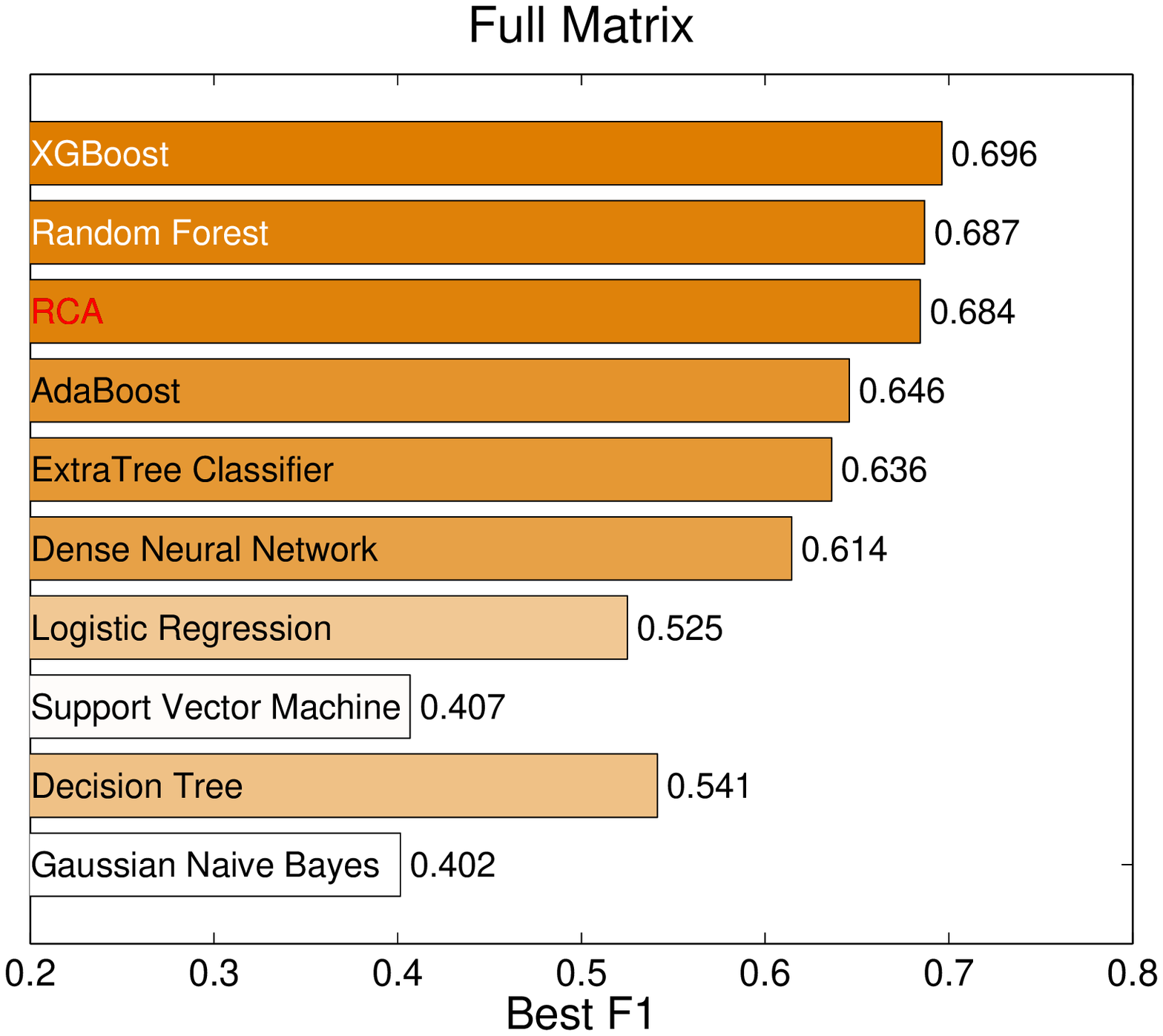}
\includegraphics[width=0.42\linewidth]{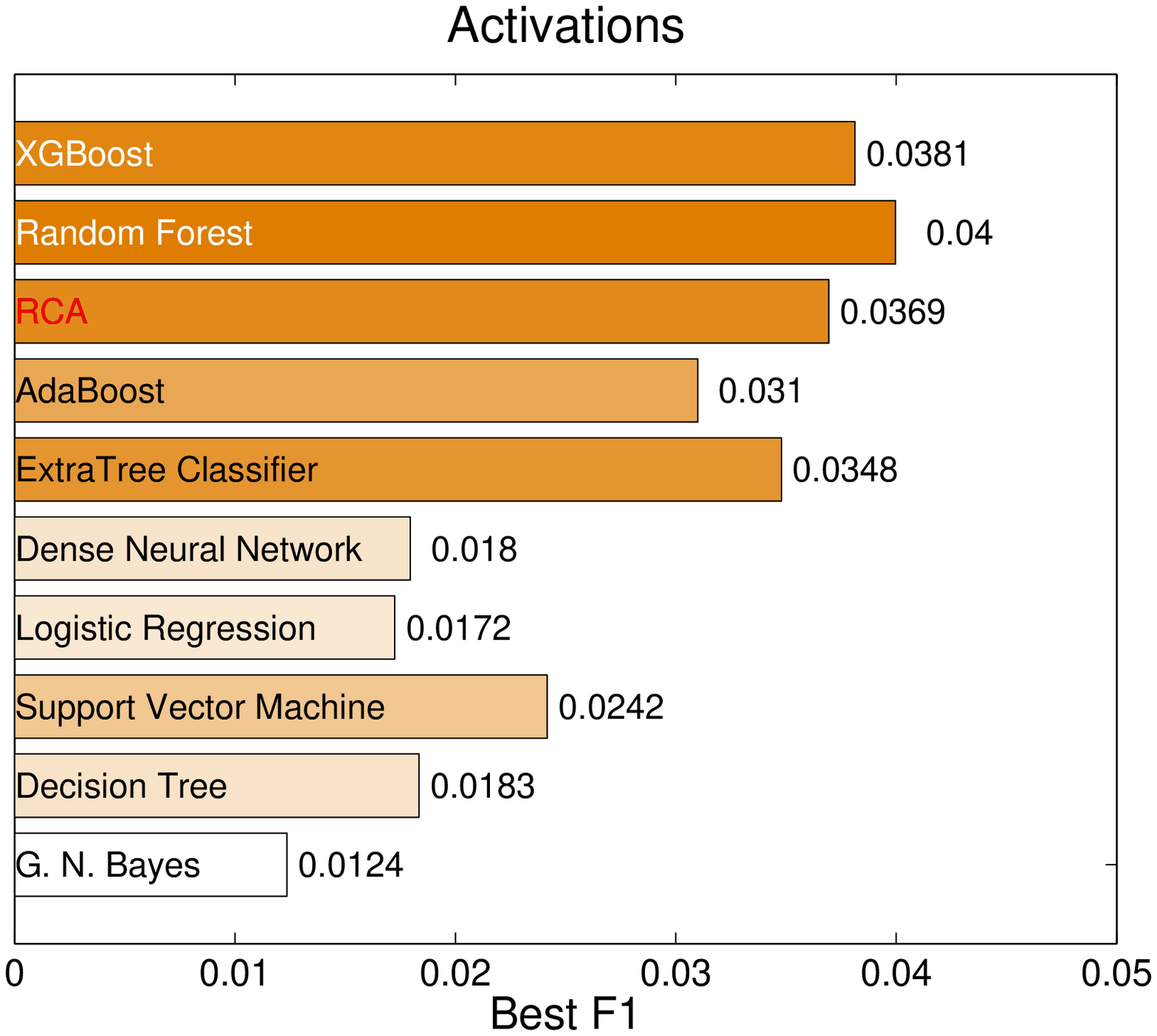}
\includegraphics[width=0.42\linewidth]{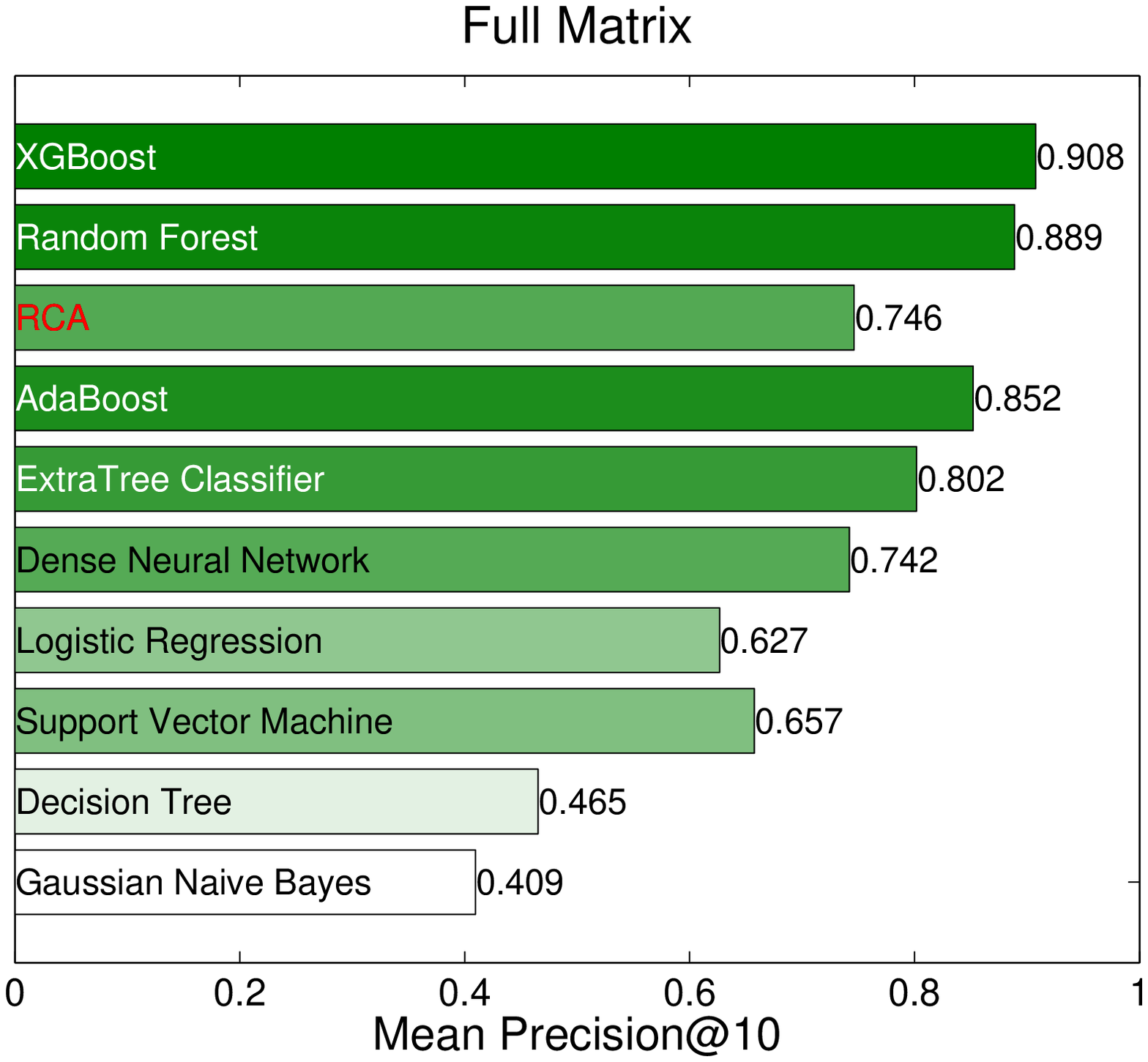}
\includegraphics[width=0.42\linewidth]{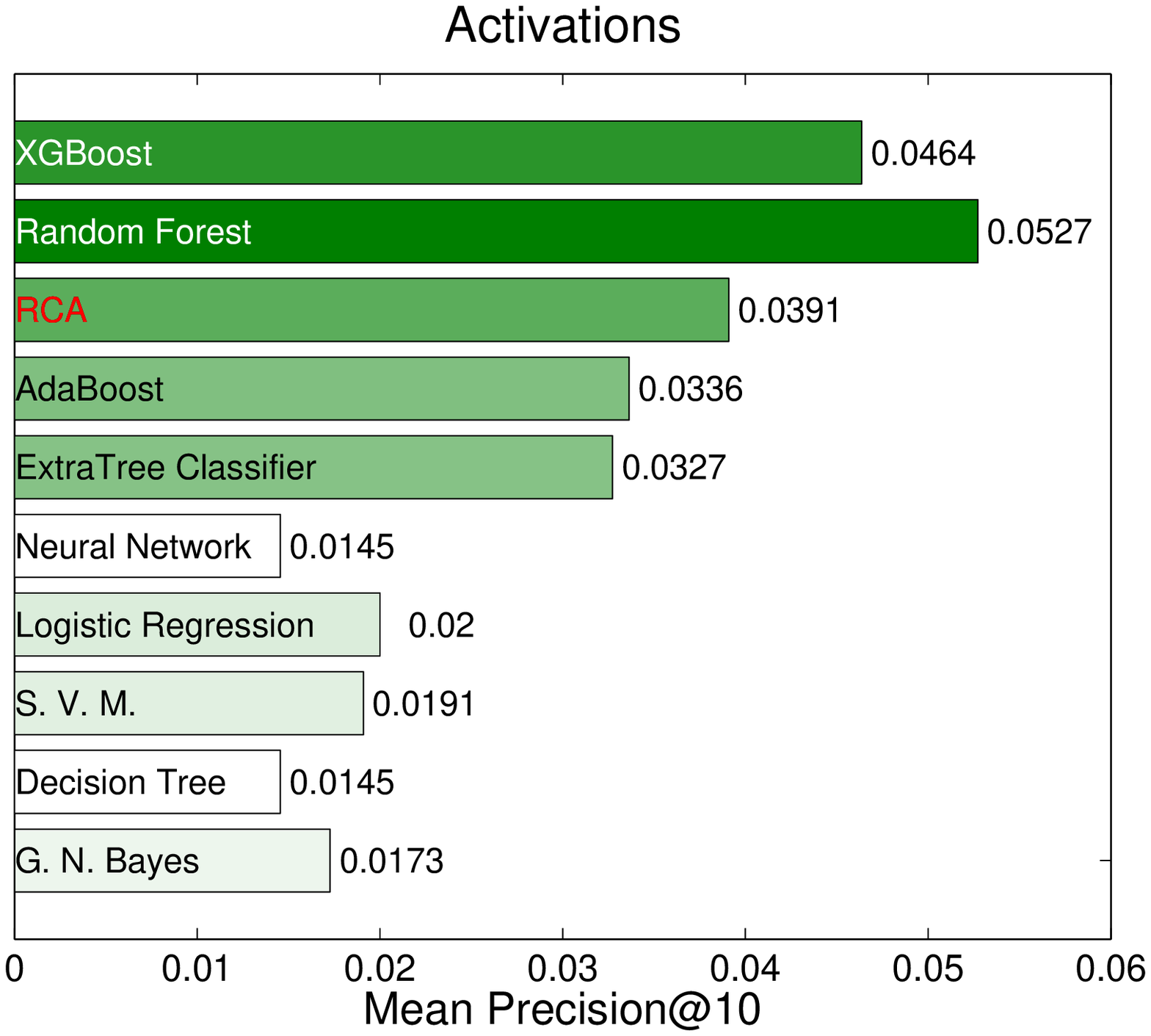}
\caption{Comparison of the prediction performance of different algorithms using three performance indicators. Tree-based approaches are performing better; the prediction of the activations is a harder task with respect to the simple future presence of a product.}
\label{fig:noCV}
\end{figure}
In order to quantitatively assess the goodness of the prediction algorithms, a number of performance indicators are available from the machine learning literature of binary classifiers. Here we focus on three of them, and we show the results in Fig.\ref{fig:noCV}, where we show a different indicator in each row, while the two columns refer to the two prediction tasks, \textit{full matrix} (i.e., the presence of a matrix element equal to one) and \textit{activations} (a zero matrix element, with RCA below 0.25 in previous years, possibly becoming higher than one, that is the appearance of a new product in the export basket of a country). AUC-ROC \cite{hanley1982meaning} is one of the most used indicators in the literature, since gives a comprehensive assessment of the prediction performance when all the predictions are ranked from the most to the less probable. The F1 Score \cite{dice1945measures,van1974foundation} is an harmonic mean of the Precision and Recall measures \cite{powers2011evaluation}, and so takes into account both False Positives and False Negatives. Finally, Mean Precision@10 considers each country separately and computes how many products, on average, are actually exported out of the top 10 predicted. All the indicators we used are discussed more in detail in the Methods section.\\ We highlight with a red color the RCA benchmark model which simply uses the RCA values in 2013 to predict the export matrix in 2018. From the analysis of Fig.\ref{fig:noCV} we can infer the following points:
\begin{enumerate}
    \item The performance indicators are higher for the full matrix. This means that predicting the unconditional presence of a product in the export basket of a country is a relatively simple task, being driven by the strong persistency of the \textbf{M} matrices through the years.
    \item On the contrary, the performance on the activations is relatively poor: for instance, on average, less than one new product of out the top ten is correctly predicted.
    \item The ROC-AUC values are very high for most algorithms in the full matrix case; this result is however biased by the strong class imbalance, as usually happens when this indicator is used \cite{fernandez2018learning}.
    \item Algorithms based on ensembles of trees, and in particular Random Forest and XGBoost, perform better than the benchmark and the the other algorithms on all the indicators.
    \item Thanks to the strong auto-correlation of the matrices the RCA-based prediction represents a very strong benchmark, also in the activations case. However, Random Forest and XGBoost perform better both in the full matrix prediction task and in the activations prediction task.
\end{enumerate}
We speculate that the machine learning algorithms perform much better in the full matrix case because, in a sense, they \textit{recognize} the single country and, when inputted with a similar export basket, they correctly reproduce the strong auto-correlation of the export matrices. We can deduce that using this approach we are not learning the complex inter-dependencies among products, as we should, and, as a consequence, we do not correctly predict the new products. In order to overcome this issue, we have to use a $k$-fold Cross Validation (CV): we separately train our models to predict the outcome of $k$ countries using the remaining $C-k$, where in our case $C=169$ and $k=13$. In this way, we prevent the algorithm to recognize the country, since the learning is performed on disjoint sets; as a consequence, the algorithm learns the relations among the products and is expected to improve the performances on the activations.\\
\begin{figure}[ht!]
\centering
\includegraphics[width=0.45\linewidth]{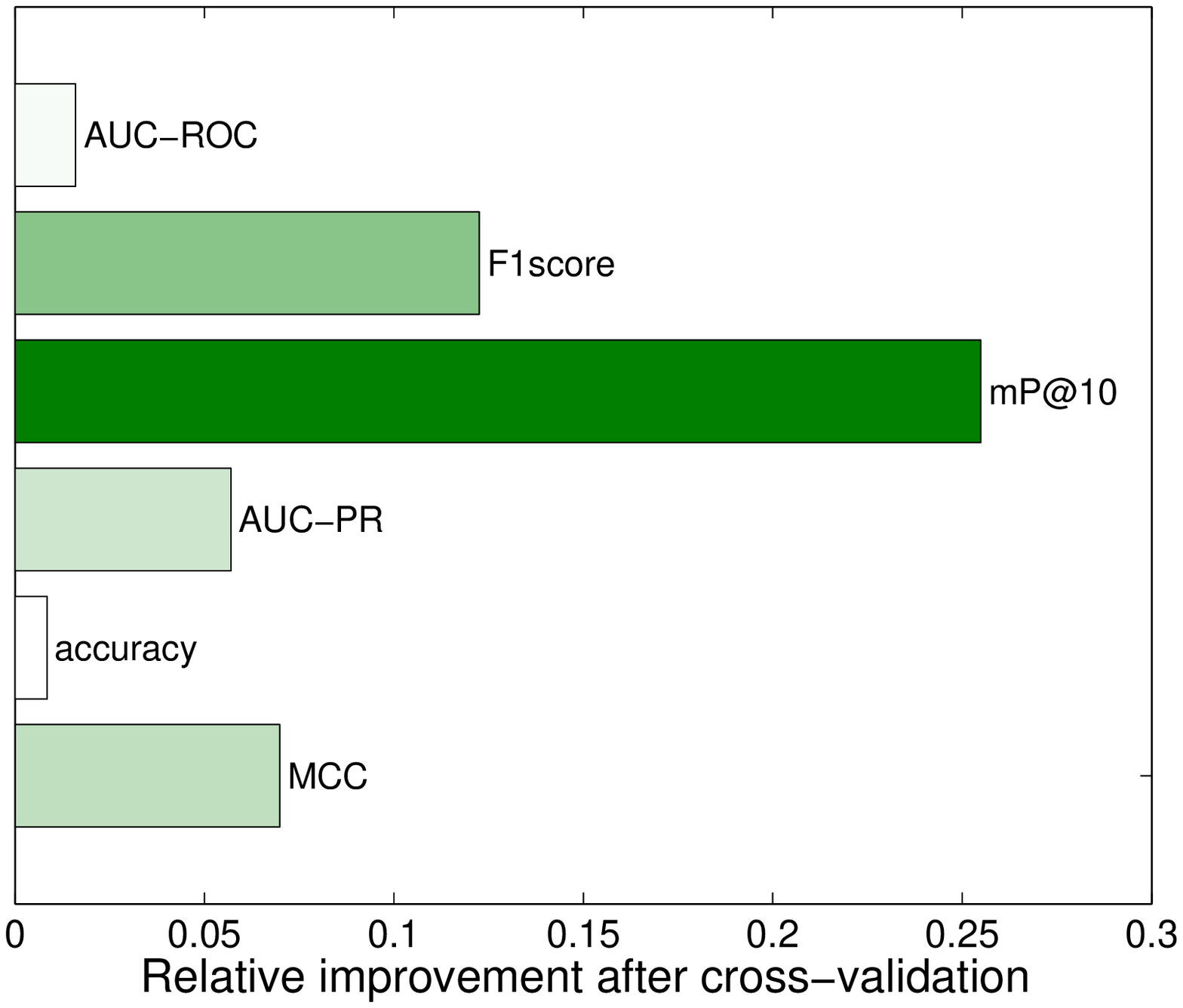}
\includegraphics[width=0.45\linewidth]{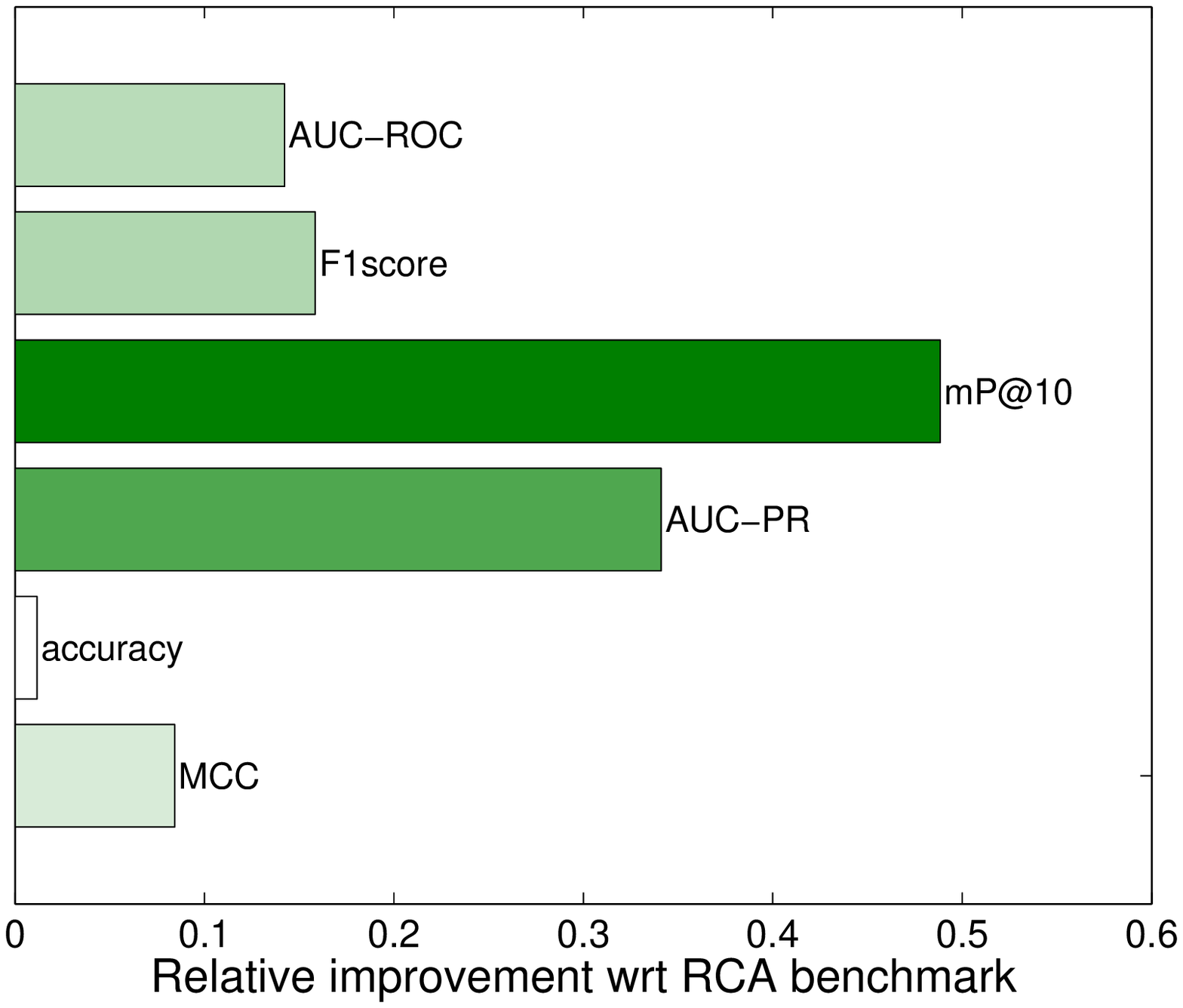}
\caption{Left: Relative improvement of the prediction performance of XGBoost when the training is cross validated. The algorithm now can not recognize the country, and so all the performance indicators improve. Right: Relative improvement of the cross validated XGBoost algorithm with respect to the RCA benchmark.}
\label{fig:siCV}
\end{figure}
Using the cross validation procedure, we trained again the two best performing algorithms which are the Random Forest and XGBoost. The result is that only the XGBoost algorithm improved its scores. The random forest gets worse and this is probably due to the fact that it is less capable than XGBoost in learning the inter-dependencies among products and it loses prediction power because with the cross validation procedure we actually use less data in the training set. So what is happening is that, if we don't do the cross validation, the random forest tends to recognize the countries better than xgboost, but if we do the cross validation xgboost learns the inter-dependencies among products better than the random forest.\\
In Fig.\ref{fig:siCV} (left) we show the relative improvements of various performance indicators when the CV is used to train the XGBoost model and the test is performed on the activations. All indicators improve; in particular, F1-score and mean Precision@10 increase of more than 10\%. On the right we compare the cross-validated XGBoost predictions with the RCA benchmark, showing a remarkable performance although the previously noted goodness of the benchmark.\\ 
In Table \ref{tab:algos} we report the values of the performance indicators for the non cross-validated Random Forest, the cross-validated XGBoost and the RCA benchmark model, one again tested on the activations. The last four rows represent the confusion matrix, where the threshold on the prediction scores is computed by optimizing the F1 scores.\\
The cross validated XGBoost gives the best scores for all the indicators, however the non cross-validated Random Forest is comparable and in any case shows better scores than the RCA model, so it represents a good alternative, especially because of the much lower relative computational cost. Indeed, the inclusion of the cross-validation procedure increases the computational cost of about a factor 13, moreover, if compared with the same number of trees, Random Forest is 7.7 times faster than XGBoost. So, even if the cross validated XGBoost model is the best performing, the non cross validated Random Forest is a good compromise to have good predictions in less time.

\begin{figure}[h!]
\centering
\includegraphics[width=0.45\linewidth]{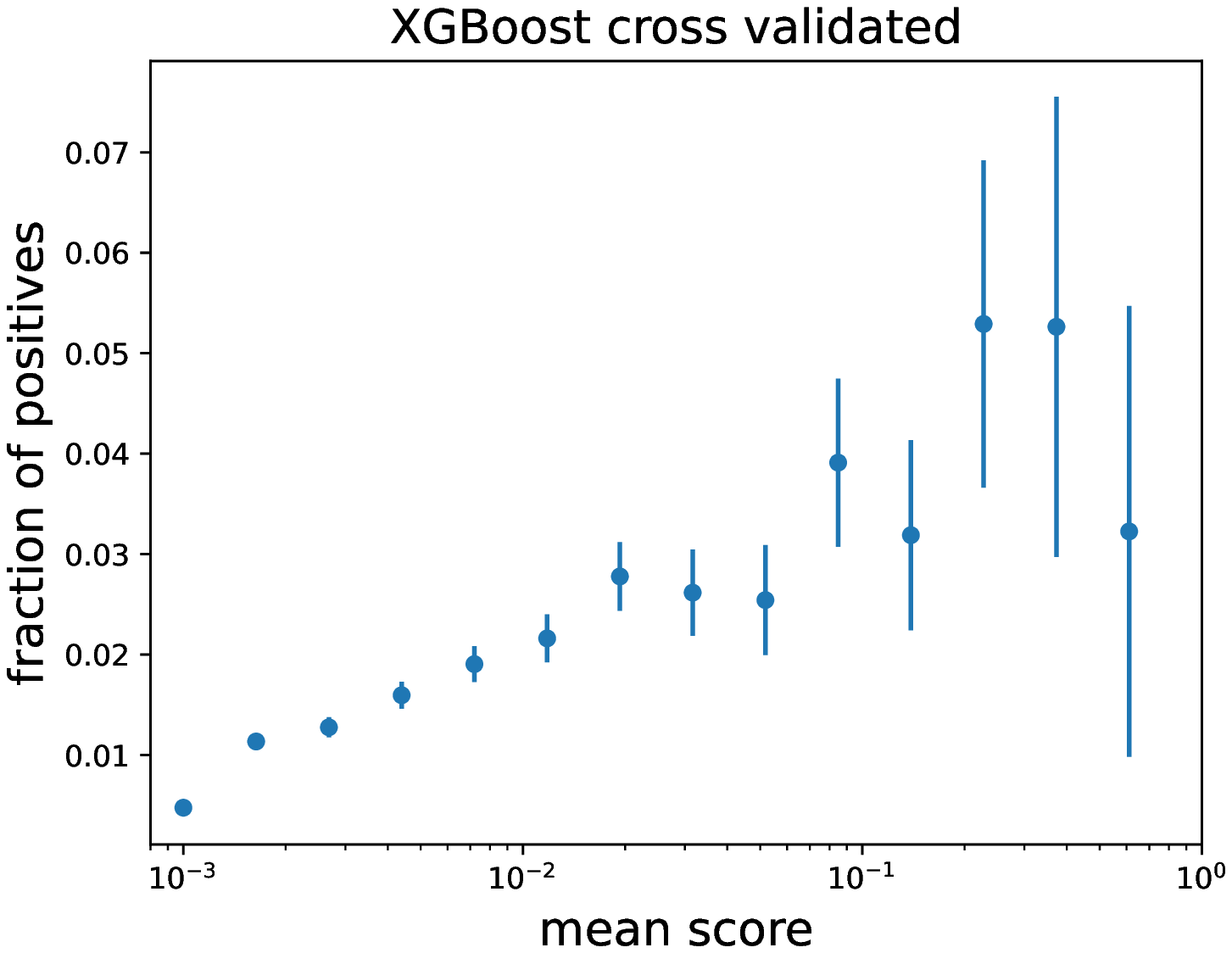}
\includegraphics[width=0.45\linewidth]{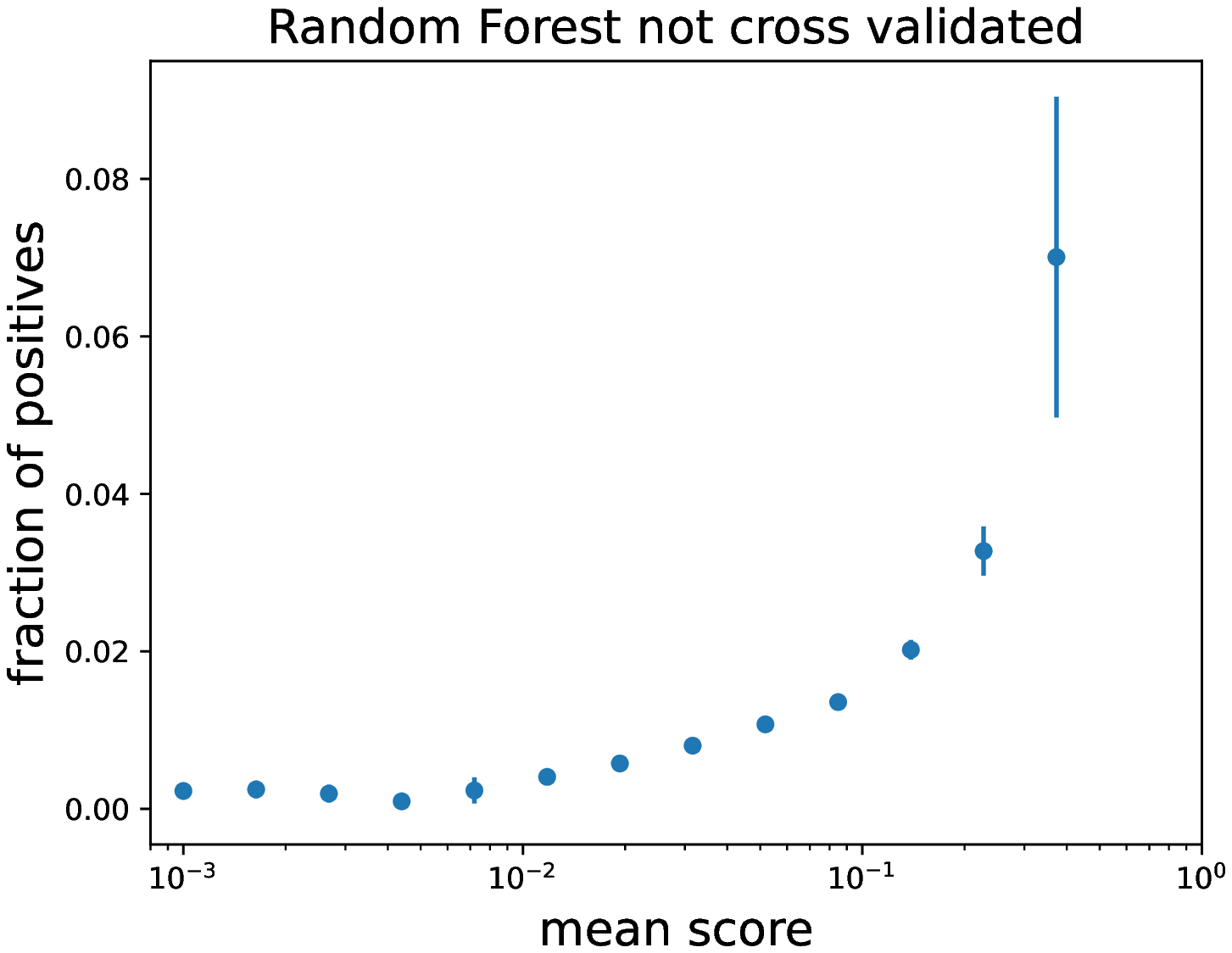}
\caption{Calibration curves: fraction of positive elements as a function of the scores produced by XGBoost (left) and Random Forest (right) for the activations prediction task. In both cases a clear positive correlation is present, indicating that higher scores are associated to higher empirical probabilities that the activation will actually occur.}
\label{fig:calibration_curve}
\end{figure}

\begin{table}[ht]
\centering
\begin{tabular}{|c|c|c|c|}
\hline
\textbf{Algorithm}        & \textbf{XGBoost} & \textbf{Random Forest} & \textbf{RCA}\\ \hline
AUC-ROC                   & 0.676            & 0.697            & 0.592                  \\ \hline
F1 score                  & 0.043            & 0.039            & 0.037                  \\ \hline
precision@10              & 0.058            & 0.053            & 0.039                  \\ \hline
precision                 & 0.029            & 0.026            & 0.023                  \\ \hline
recall                    & 0.079            & 0.083            & 0.103                  \\ \hline
MCC                       & 0.038            & 0.036            & 0.035                  \\ \hline
AUC-PR                    & 0.015            & 0.015            & 0.011                  \\ \hline
accuracy                  & 0.978            & 0.975            & 0.967                  \\ \hline
negative predictive value & 0.994            & 0.994            & 0.994                  \\ \hline
TP                        & 203             & 213            & 263                   \\ \hline
FP                        & 6722            & 8079            & 11413                  \\ \hline
FN                        & 2358            & 2348            & 2298                  \\ \hline
TN                        & 402708           & 401351            & 398017                 \\ \hline
computational cost        & ~100            & 1                 & -     \\ \hline
\end{tabular}
\caption{Predictive performance of XGBoost with cross validation, Random Forest without cross validation and the RCA benchmark for the activations. The last row indicates the computational cost with respect the non cross validated Random Forest; XGBoost is about 100 times slower.}
\label{tab:algos}
\end{table}
In general, a desiderable output of a classification task is not only a correct prediction, but also an assessment of the likelihood of the label, in this case, the activation. This likelihood provides a sort of confidence on the prediction. In order to test whether the scores are correlated or not with the actual probability of activations we build a calibration curve. In Fig. \ref{fig:calibration_curve} we show the fraction of positive elements as a function of the output (i.e., the scores) of the XGBoost and Random Forest algorithms in the activations prediction task. We divide the scores in logarithmic bins and then we compute mean and standard deviation. In both cases a clear correlation is present, pointing out that a higher prediction score correspond to a higher empirical probability that the activation of a new product will actually occur. Moreover, we note that the greater is the score produced by the model, the greater is the error on the y axis; the reason is that the models tend to assign higher scores to the products already exported from a country, so if we look to the activations the values start to fluctuate, and the statistics becomes lower.

\subsection*{Opening the black box}
\begin{figure}[ht!]
\centering
\includegraphics[width=\linewidth]{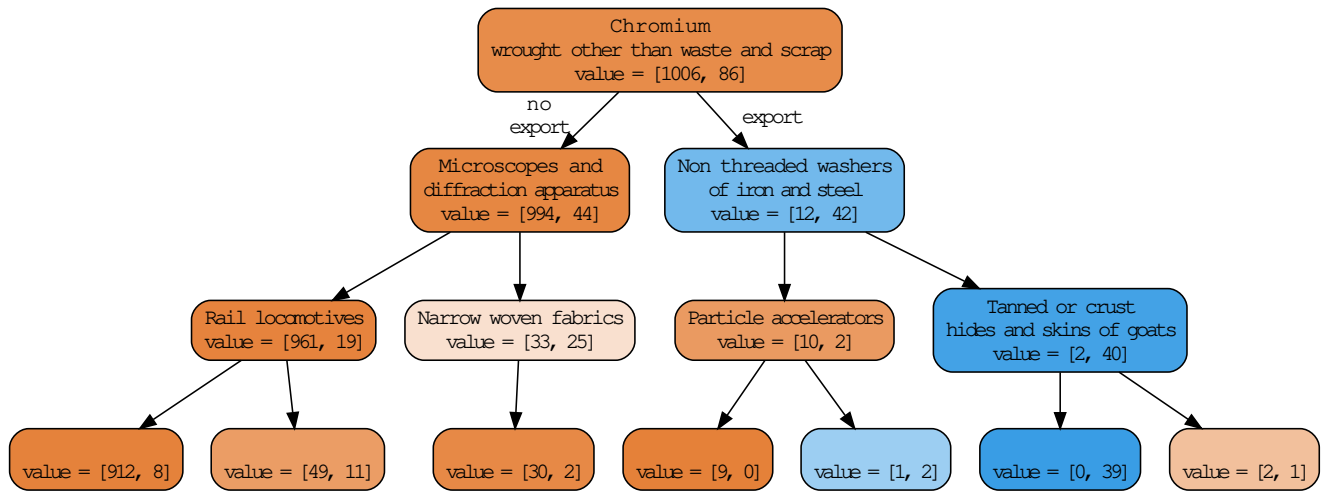}
\caption{A representative decision tree to forecast the export of the product \textit{valves and tubes}. The root product, \textit{chromium}, has a well known technological relation with the target product, and in fact is able to discriminate future exporters with high precision.}
\label{fig:tree}
\end{figure}

In order to qualitatively motivate the better performance of tree-based algorithms, in this paragraph we elaborate on the operation of Random Forests. As specified in the Methods section, in these prediction exercises we train one Random Forest model for each product, and each Random Forest contains 100 decision trees. In Fig.\ref{fig:tree} we show one representative decision tree. This tree is obtained by putting the number of features available for each tree equal to $P=5040$: this means that we are bootstrap aggregating, or \textit{bagging}\cite{breiman1996bagging} the trees, instead of building an actual random forest, which considers instead a random subset of the products\cite{breiman2001random} (the decision trees may be different also in this case, since the bagging procedure extracts the features with replacement). Moreover, the training procedure is cross validated, so the number of input countries is 156$\times$7 (156 countries and 7 years from 2007 to 2013).\\
The decision tree we show refers to the product with HS1992 code 854089; the description is \textit{valves and tubes not elsewhere classified in heading no. 8540}, where 8540 stands for \textit{cold cathode or photo-cathode valves and tubes like vacuum tubes, cathode-ray tubes and similars}.\\
The color represents the class imbalance of the leaf (dark orange, many zeros; dark blue, many ones, quantified in the square brackets). The root product, the one which provides the best split, is \textit{chromium}, which is used, for instance, in the cathode-ray tubes to reduce X-ray leaks. So the random forest found a non trivial connection between chromium and these types of valves and tubes: out of the 1006 couples country-year that do not export valves and tubes, 994 do not export chromium either (note the negative association). We can explore the network considering that the no-export link is always on the left. Looking to the export direction we find the cut on washers of iron and steel that works very well: only 2 of the 12 couples country-year that do not export valves and tubes do export washers and only 2 of the 42 countries that export valves and tubes do not export washers.\\
Looking to the other splits we find some of them more reasonable, like the one on particle accelerators, and some that seem coincidental, like the one on hides and sinks of goat.\\
From this example it is clear that the decision tree is a natural framework to deal with a set of data in which some features (i.e., products) may be by far more informative than others, and so a hierarchical structure is needed to take into account this heterogeneous feature importance.\\
\begin{figure}[ht!]
\centering
\includegraphics[width=0.7\linewidth]{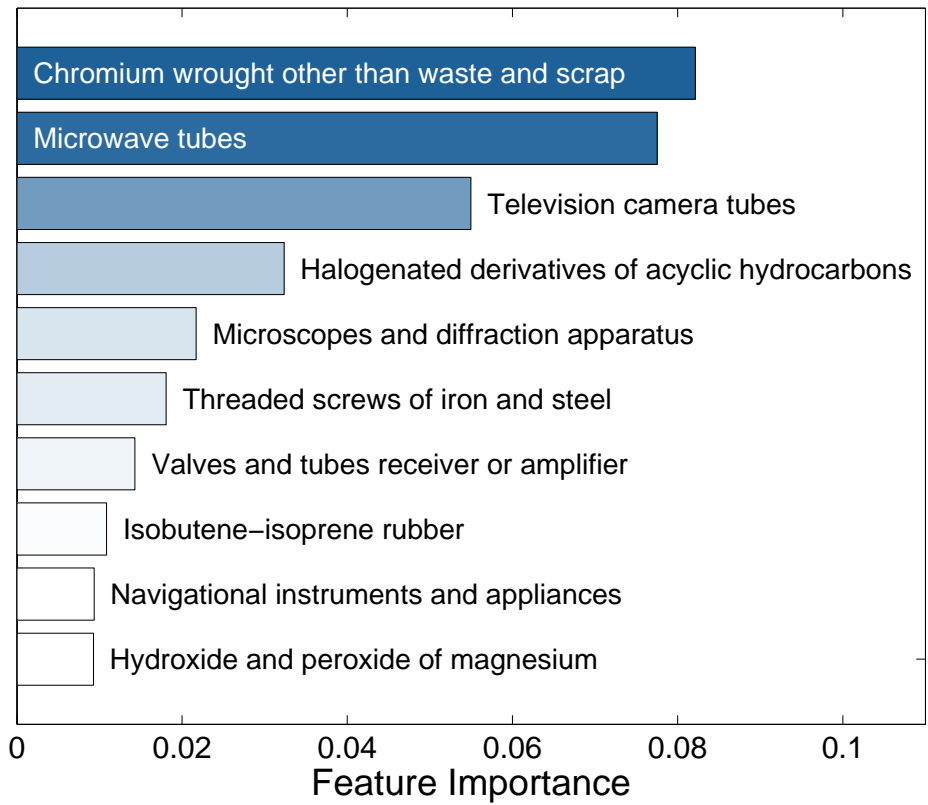}
\caption{Feature importance is a measure of how much a product is useful to predict the activation of the target product. Here we use the average reduction of the Gini impurity at each split. All important products are reasonably connected with the target.}
\label{fig:featimp}
\end{figure}
Feature importance may be evaluated by looking at the normalized average reduction of the impurity at each split that involves that feature \cite{pedregosa2011scikit}. In our case, we are considering the Gini impurity. In Fig.\ref{fig:featimp} we plot this assessment of the feature importance to predict the activation of valves and tubes. One can easily see that the average over the different decision trees is even more meaningful than the single decision tree shown before, even if the each one of former see less products of the latter: all the top products are reasonably connected with the target product and so it is natural to expect them to be key elements to decide whether the given country will export valves and tubes or not.
\subsection*{Time dependence}
In the procedure discussed above we used a time interval $\Delta_{model}$ equal to 5 years for the training, and we tested our out-of-sample forecasts using the same time interval $\Delta$. Here we investigate how the choice of the forecast horizon $\Delta$ affects the quality of the predictions. To make this analysis we used XGBoost models trained with the cross validation method a lower $\Delta_{model}$ = 3. The machine learning algorithms are trained using data in the range $y \in [1996\dots2008]$ and their output, obtained giving \textbf{RCA}$^{(2008)}$ as input, will be compared with the various \textbf{M}$^{(2008+\Delta)}$ by varying $\Delta$. Being the last year of available data 2018, we can explore a range of $\Delta$s from 1 to 10. All details about the training procedure of the machine learning algorithms are given in the Methods section.\\
\begin{figure}[ht!]
\centering
\includegraphics[width=0.48\linewidth]{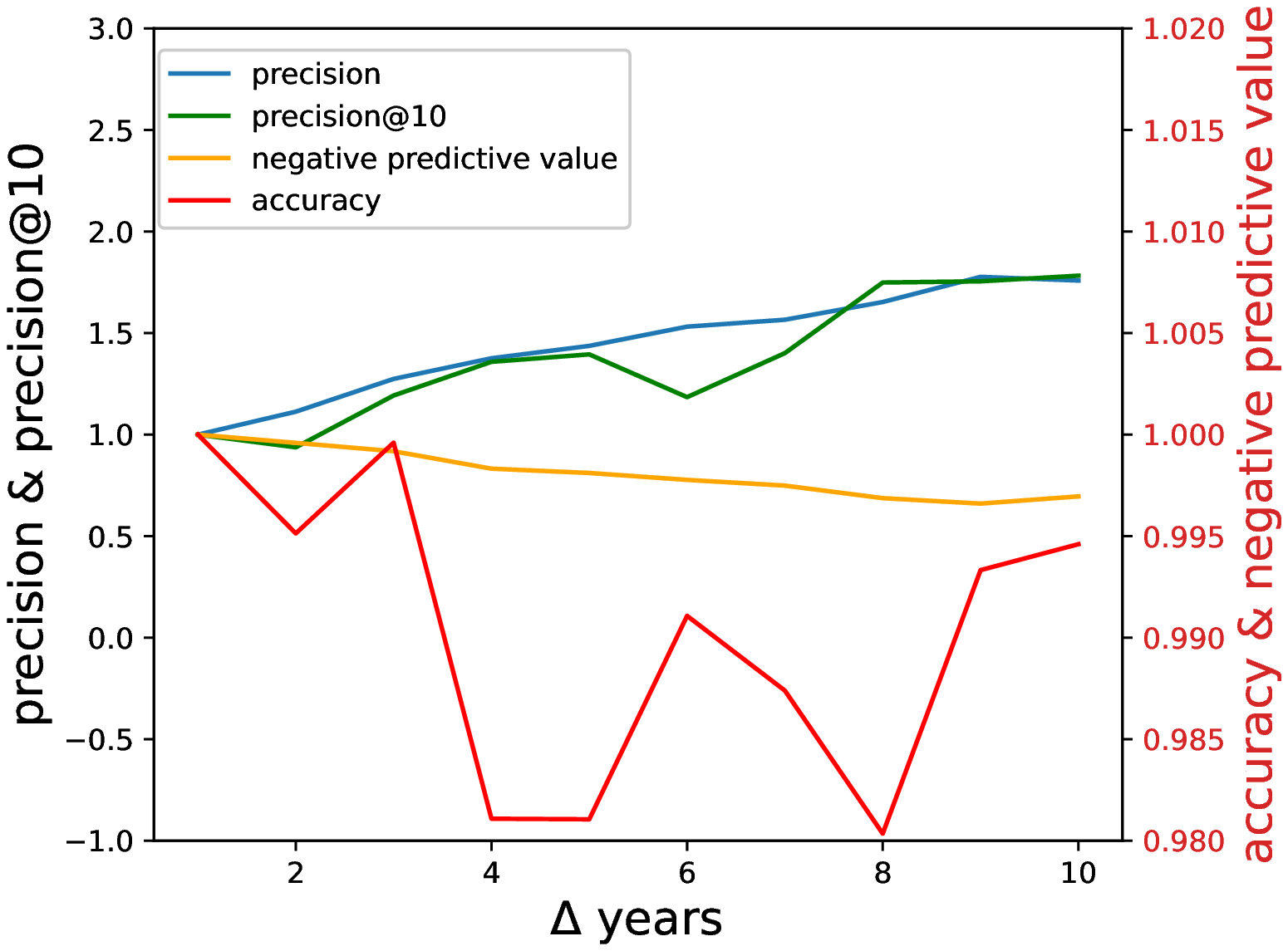}
\includegraphics[width=0.48\linewidth]{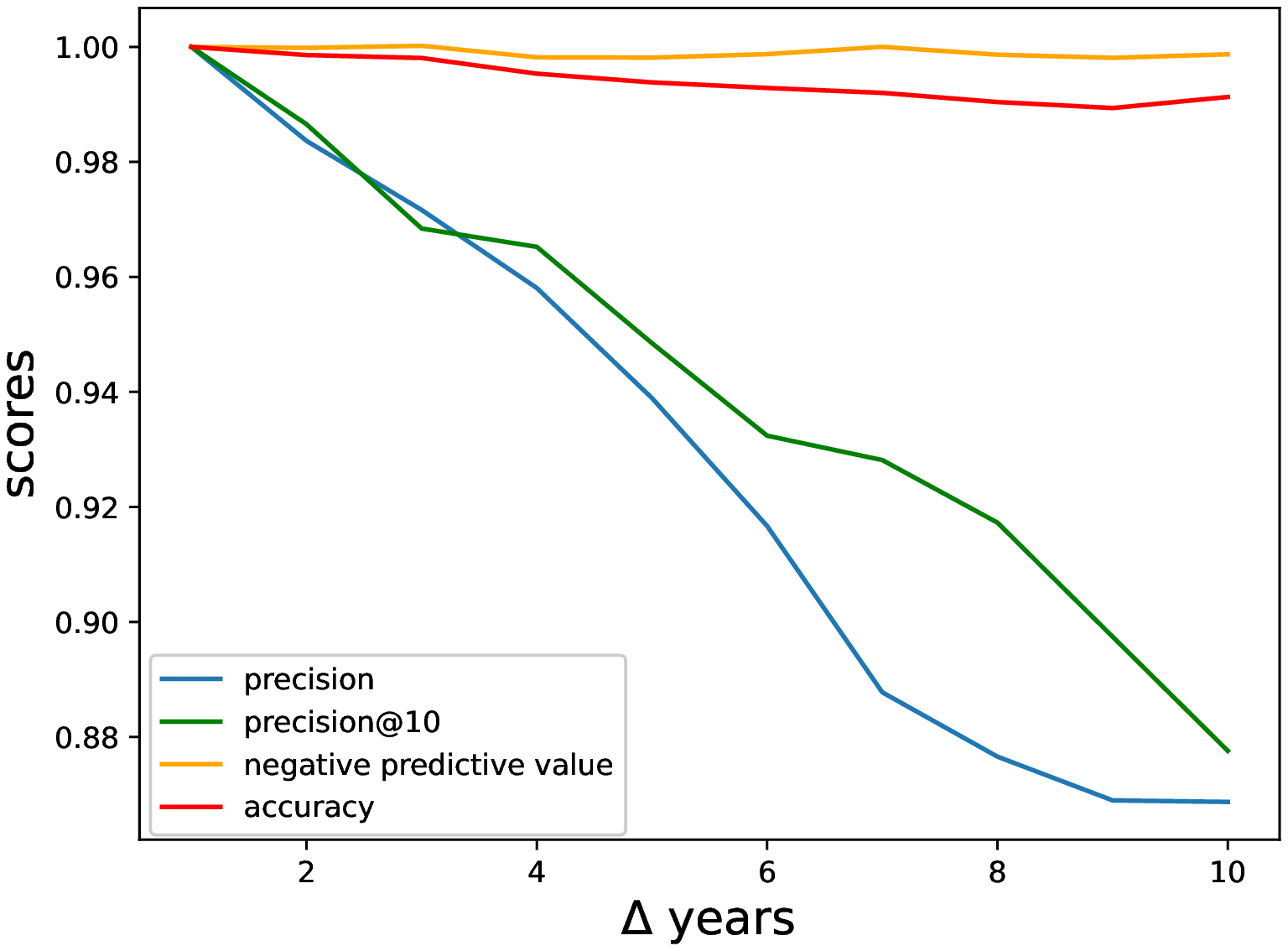}
\caption{In the plot on the left we show the performance indicators in the case of the activations prediction task. The performance on positive values improves, while the one on negative values gets worse. On the right we show the same performance indicators in the case of the full matrix prediction task. All the scores gets worse due to the vanishing auto-correlation of the matrices.}
\label{fig:delta_analysis}
\end{figure}
The quality of the predictions as a function of the forecast horizon $\Delta$ are summarized in Fig.\ref{fig:delta_analysis}, where we normalized the indicators in such a way that they are all equal to 1 at $\Delta=1$. In the left figure we have the plot for the activations prediction task: both \textit{precision} and \textit{precision@10} increase with $\Delta$, while the \textit{negative predictive value} decreases and accuracy shows an erratic behavior. This means that our ability to guess positive values improves or, in other words, the greater is the time you wait the higher is the probability that a country sooner or later does activate the products we predict. This improvement on positive values, however, corresponds to a worsening on negative values that can be interpreted as the fact that countries during time develop new capabilities and start to export products we cannot predict with a $\Delta$ interval too large.\\
If we look to a score that includes both performance on positive values and performance on negative values, like accuracy, we have a (noisy) worsening with the increase of $\Delta$.\\
In the figure on the right we show instead the full matrix prediction task. In this case all the scores decrease with $\Delta$ because the algorithm can not leverage anymore on the strong auto-correlation of the RCA matrix.\\
Note that the steepness of the decreasing curves is higher when we look at precision scores, the reason being the high class imbalance and the large amount of true negatives with respect true positives as you can see in Table \ref{tab:algos}.

\section*{Discussion}
One of the key issues in economic complexity and, more in general, in complexity science is the lack of systematic procedures to test, validate, and falsify theoretical models and empirical, data-driven methodologies. In this paper we focus on export data, and in particular on the country-product bipartite network, which is the basis of most literature in economic complexity, and the likewise widespread concept of \textit{relatedness}, that is usually associated to an assessment of the proximity between two products or the density or closeness of a country with respect to a target product. As detailed in the Introduction, many competing approaches exist to quantify these concepts, however, a systematic framework to evaluate which approach works better is lacking, and the result is the flourishing of different methodologies, each one tested in a different way and with different purposes. We believe that this situation can be discussed in a quantitative and scientifically sound way by defining a concrete framework to compare the different approaches in a systematic way; the framework we propose is out-of-sample forecast, and in particular the prediction of the presence or the appearance of products in the export baskets of countries. This approach has the immediate advantage to avoid a number of recognized issues \cite{romer2016trouble} such as the mathiness of microfounded models \cite{romer2015mathiness} and the p-hacking in causal inference and regression analyses \cite{athey2018impact,head2015extent}.\\
In this paper we systematically compare different machine learning algorithms in the framework of a supervised classification task. We find that the statistical properties of the export data, namely the strong auto-correlation and the class imbalance, imply that the appearance, or activation, of new products should be investigated, and some indicators of performance, such as ROC-AUC and accuracy, should be considered with extreme care. On the contrary, indicators such as the mean Precision@k are much more stable and have an immediate policy interpretation. 
In the prediction tasks tree-based models, such as Random Forest and Boosted Trees, clearly overperform the other algorithms and the quite strong benchmark provided by the simple RCA measure. The prediction performance of Boosted Trees can be further improved by training them in a cross validation setting, at the cost of a higher computational effort. The calibration curves, which show a high positive correlation between the machine learning scores and the actual probability of the activation of a new product, provide further support to the correctness of these approaches. A first step towards opening this black box is provided by the visual inspection of a sample decision tree and the feature importance analysis, that shows that the hierarchical organization of the decision tree is a key element to provide correct predictions but also insights about which products are more useful in this forecasting task.\\
From a theoretical perspective, this exercise points out the relevance of context for the appearance of new products, in the spirit of New Structural Economics \cite{lin2012new}, but this has also immediate policy implications: each country comes with its own endowments and should follow a personalized path, and machine learning approaches are able to efficiently extract this information. In particular, the output of the Random Forest or the Boosted Trees algorithm, provide scores, or \textit{progression probabilities} that a product will be soon activated by the given country. This represents a quantitative and scientifically tested measure of the \textit{feasibility} of a product in a country. This measure can be used in very practical contexts of investment design and industrial planning, a key issue after the covid-related economic crisis \cite{fernandes2020economic,nana2020when}.\\
This paper opens up a number of research lines in various directions. We are planning to compare these results with machine learning algorithms and network-based approaches explicitly designed for link prediction tasks \cite{al2006link,medo2018link}, to apply this framework to different bipartite networks using different databases, and to use statistically validated projections \cite{pugliese2019unfolding} to build density-based predictions. All these studies will be presented in future works.
\section*{Methods}
\subsection*{Data description}
The data we use in this analysis are obtained from the UN-COMTRADE database, Harmonized System 1992 classification (HS 1992) and includes the volumes of the export flows between countries. The time range covered is 1996-2018 and for each year we have a matrix $\textbf{V}$ whose element $V_{cp}$ is the amount, expressed in US dollars, of product p exported by country c. The total number of countries is 169 and the total number of products is 5040.\\
To binarize the data we determine if a country competitively exports a product by computing the Relative Comparative Advantage (RCA) introduced by Balassa \cite{balassa1965trade}. The RCA of a country c in product p in year y is given by:
\begin{equation}
R^{(y)}_{cp} = \left. \frac{V^{(y)}_{cp}}{\sum_{p'}V^{(y)}_{cp'}}   \middle/ \frac{\sum_{c'}V^{(y)}_{c'p}}{\sum_{c'p'}V^{(y)}_{c'p'}} \right.
\end{equation}
$R^{(y)}_{cp}$ is a continuous value and represents the ratio between the weight of product p in the export basket of country c and the total weight of that product in the international trade. Alternatively, the RCA can be seen as the ration between the market share of country c relatively to product p and the weight of country c with respect to the total international trade. This is the standard way, in the economic complexity literature, to remove trivial effects due to the size of the country and the size of the total market of the product. In this way, a natural threshold of one can be used to establish whether country c exports product p in a competitive way or not. As a consequence, the last matrix we need to define in our analysis is the matrix $\textbf{M}$ whose binary element $M_{cp}$ tells us if country c is competitive in the export of product p or not:
\begin{equation}
M^{(y)}_{cp} = 
\left \{ \begin{array}{rl}
1~~~~if~~R_{cp}^{(y)} \geq 1\\
0~~~~if~~R_{cp}^{(y)} < 1
\end{array}
\right.
\end{equation}
In this work we will try to predict future values of $M_{cp}$ using past values of RCA.
\subsection*{Training and testing procedure}
\begin{figure}[ht!]
\centering
\includegraphics[width=0.48\linewidth]{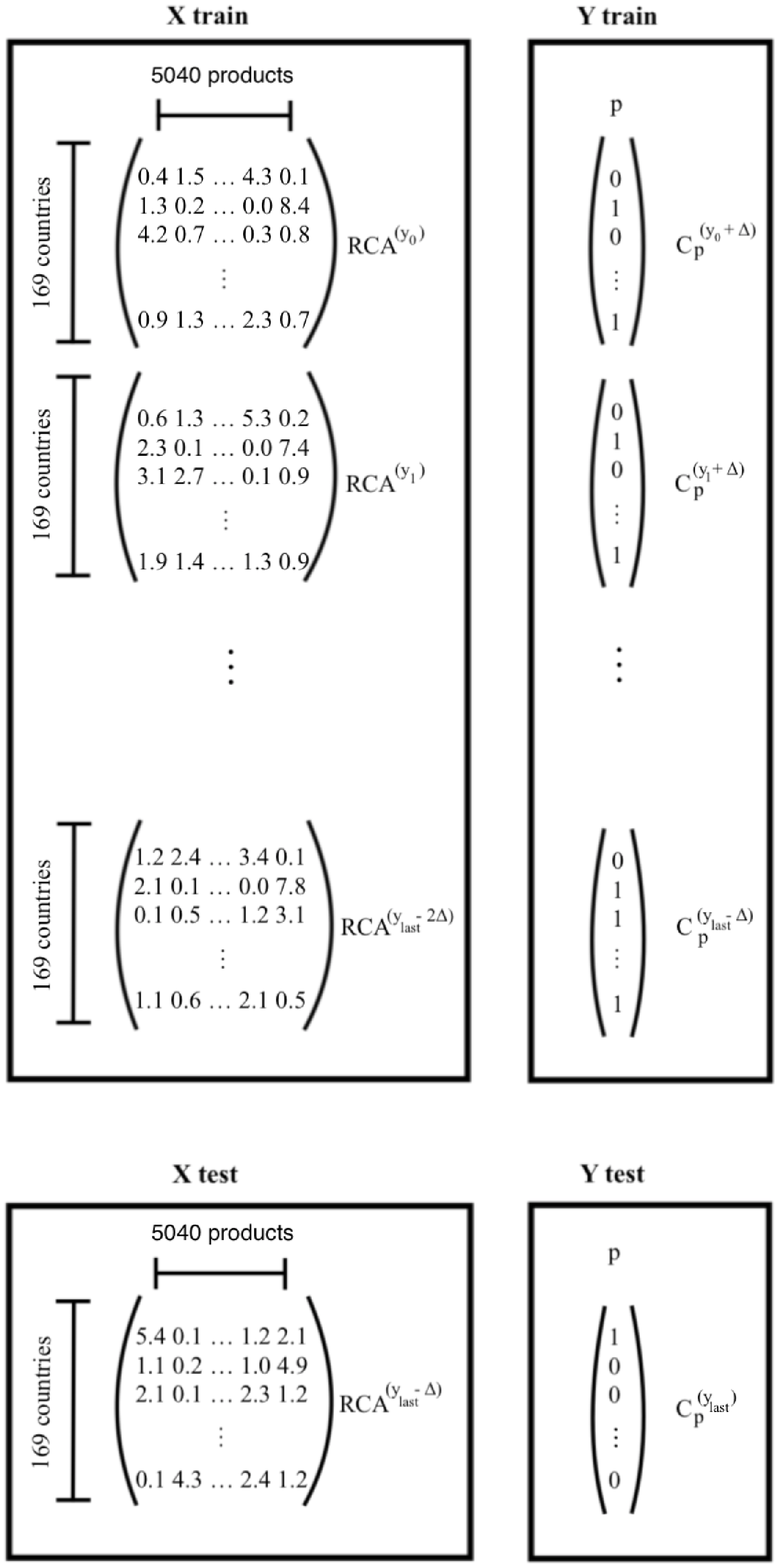}
\includegraphics[width=0.48\linewidth]{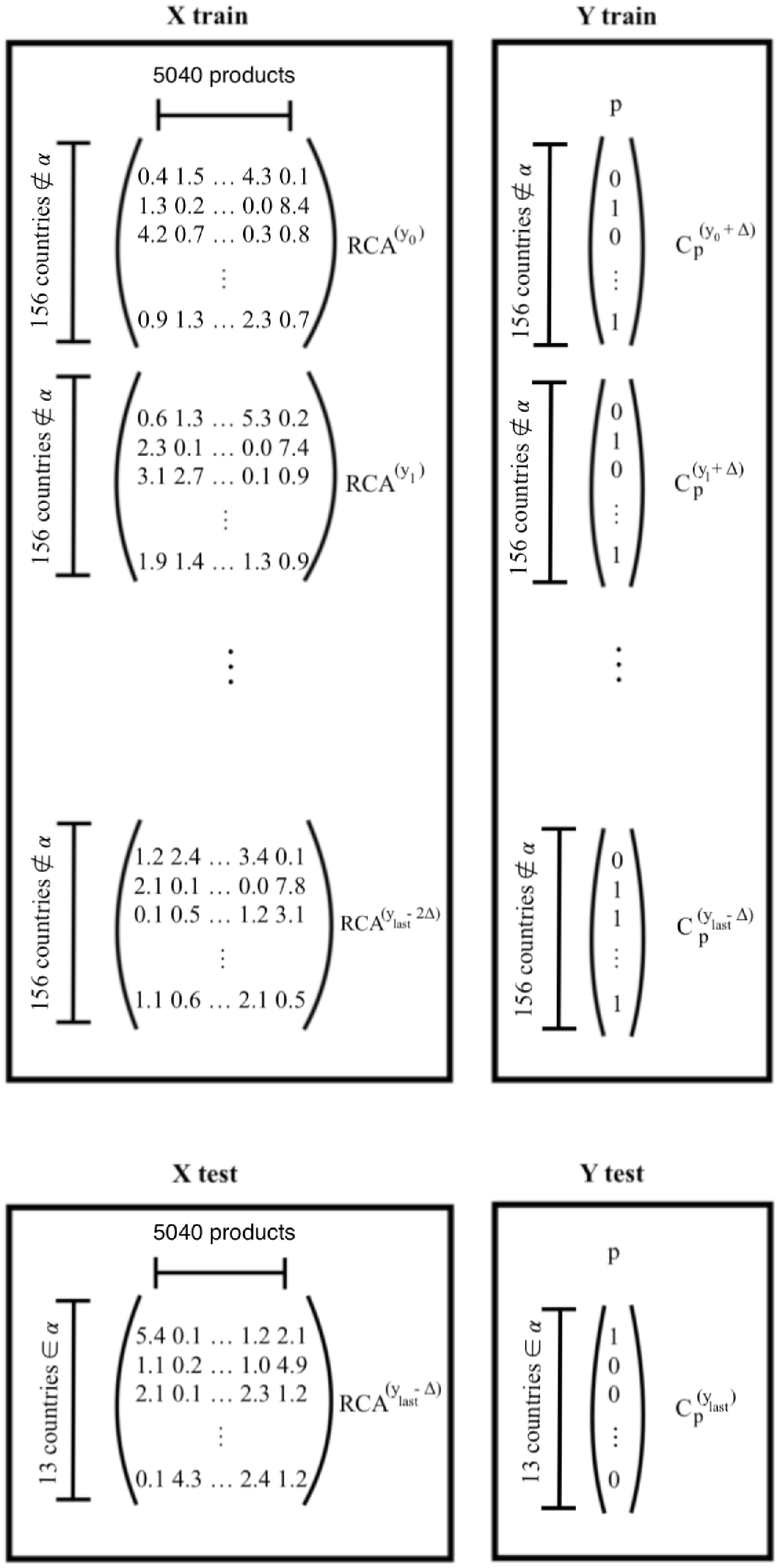}
\caption{The training and testing procedure with (right) and without (left) cross validation. See the text for a detailed explanation.}
\label{fig:schemini}
\end{figure}

We want to guess which products will be exported by a country after $\Delta$ years. To do this, we exploit machine learning algorithms with the goal to (implicitly) understand the capabilities needed to export a product from the analysis of the export basket of countries. Since each product requires a different set of capabilities, we need to train different models: for instance, we have to train 5040 different random forests, one for each product.\\
The training procedure is analogous for all the models: they have to connect the RCA values of the products exported by a country in year y with the element $M^{(y+\Delta)}_{cp}$, which tells us if country c in year $y+\Delta$ is competitive in the export of product p.\\
In the general case we have export data that covers a range of years [$y_0$, $y_{last}$]. The last year is used for the test of the model and so the training set is built using only the years [$y_0$, $y_{last}-\Delta$]. In this way, no information about the $\Delta$ years preceding $y_{last}$ is given.\\
The input of the training set, that we call \textbf{X}$_{train}$, is vertical stack of the \textbf{R}$^{(y)}$ matrices from $y_0$ to $y_{last}-2\Delta$ (see Fig.\ref{fig:schemini}). In such a way we can consider all countries and all years of the training set, and these export baskets will be compared with the corresponding presence or absence of the target product p after $\Delta$ years; this because our machine learning procedure is supervised, that is, during the training we provide a set of answers \textbf{Y}$_{train}$ corresponding to each export basket in \textbf{X}$_{train}$. While \textbf{X}$_{train}$ is the same for all the models (even if they refer to different products), the output of the training set \textbf{Y}$_{train}$ changes on the basis of the product we want to predict. If we consider the model associated to product p, to build \textbf{Y}$_{train}$ we aggregate the columns corresponding to the target product, C$^{(y)}_p$, of the \textbf{M} matrices from $y_0+\Delta$ to $y_{last}-\Delta$ (so we use the same number of years, all shifted by $\Delta$ years with respect to \textbf{X}$_{train}$). This is graphically represented on the extreme left side of fig.\ref{fig:schemini}.\\
Once the model is trained, in order to perform the test we give as input \textbf{X}$_{test}$ the matrix \textbf{R}$^{(y_{last}-\Delta)}$. Each model will give us its prediction for the column p of the matrix \textbf{M}$^{(y_{last})}$ and, putting all the results together, we reconstruct the whole matrix of scores \textbf{M}$_{pred}^{(y_{last})}$ which can be compared with the empirical one. There are various ways to compare the predictions with the actual outcomes, and these performance metrics are discussed in the following section.\\
As already mentioned, the same models can be tested against two different prediction tasks: either we can look to the full matrix \textbf{M}$^{(y_{last})}$, either we can concentrate only on the possible \textit{activations}, that is products that were not present in an export basket and countries possibly start exporting. The set of possible activations is defined as follows:
\begin{equation}
    (c,p) \in activations \iff R^{(y)}_{cp} < 0.25~~\forall y \in [y_0,y_{last}-\Delta] 
\end{equation}
In other words a pair (c,p) is a possible activation if country c has never been competitive in the export of product p till year $y_{last}-\Delta$, that is its RCA values never exceeded 0.25. This selection of the test set may look too strict, however it is key to test our algorithms against situations in which countries really start exporting new products. Because of the RCA binarization, there are numerous cases in which a country noisily oscillates around RCA=1 and, de facto, that country is already competitive in that product; in these case the RCA benchmark is more than enough for a correct prediction.\\ 
The way to train the models we just described performs better on the full matrix than in the activations. The reason is probably that the machine learning algorithms recognize the countries because the ones in the training set and the ones in the test set are the same. When the algorithms receive as input the export basket of a country they have already seen in the training data, they tend to reproduce the strong autocorrelation of the export matrices. To avoid this problem we used a k-fold cross validation, which means that we split the countries in k groups. Since the number of countries is 169, the natural choice is to use k = 13, so we randomly extract a group $\alpha$ of 13 countries from the training set, which is then composed by the remaining 156 countries, and we use only the countries contained in $\alpha$ for the test. In this way each model is meant to make predictions only on the countries of group $\alpha$, so to cover all the 169 countries we need to repeat the procedure 13 times, every time changing the countries in group $\alpha$. This different training procedure is depicted on the right part of fig.\ref{fig:schemini}. So there will be 13 models associated to a single product and, for this reason, the time required to make the training is 13 times longer. Like in the previous case, in the training set we aggregate on the years in the range [$y_0$, $y_{last}-\Delta$]. \textbf{X}$_{train}$ is the aggregation of the RCA matrices from $y_0$ to $y_{last}-2\Delta$ and \textbf{Y}$_{train}$ is the aggregation of the column p of the M matrices from $y_0+\Delta$ to $y_{last}-\Delta$. In both cases, the countries in group $\alpha$ are removed.\\
When we perform the test, each models takes as \textbf{X}$_{test}$ the matrix \textbf{RCA}$^{(y_{last}-\Delta)}$ with only the rows corresponding to the 13 countries in group $\alpha$ and gives as output scores the elements of the matrix \textbf{M}$_{pred}^{(y_{last})}$. All the 5040$\times$13 models together give as output the whole matrix of scores \textbf{M}$_{pred}^{(y_{last})}$ that will be compared to the actual \textbf{Y}$_{test}$ = \textbf{M}$^{(y_{last})}$.\\
Since the output of the machine learning algorithms is a probability, and most of the performance indicators require a binary prediction, in order to establish if we predict a value of 0 or 1 we have to introduce a threshold. The value of this threshold we use is the one that maximizes the F1-score. We note that the only performance measure that do not require a threshold are the ones that consider areas under the curves, since these curves are built precisely by varying the threshold value.\\
Fig.\ref{fig:schemini} schematically shows the training procedures with and without cross validation.

\subsection*{Performance indicators}
The choice of the performance indicators is a key issue of supervised learning\cite{caruana2006empirical,powers2011evaluation} and, in general, strongly depends on the specific problem under investigation. Here we discuss the practical meaning of the performance indicators we used to compare the ML algorithms. For all the scores but the areas under curves, we need to define a threshold above which the output scores of the ML algorithms are associated with a positive prediction. For this purpose we choose the threshold that maximizes the F1 score\cite{lipton2014optimal}. 
\begin{itemize}
\item \textbf{Precision} Precision is defined as the ratio between true positives and positives \cite{powers2011evaluation}. In our case, we predict that a number of products will be competitively exported by some countries; these are the \textit{positives}. The precision is the fraction that counts how many of these predicted products are actually exported by the respective countries after $\Delta$ years. A high value of precision is associated to a low number of false positives, that is if products that are predicted to appear they usually do so.
\item \textbf{mean Precision@k (mP@k)}
This indicator usually corresponds to the fraction of the top $k$ positives that are correctly predicted. We considered only the first $k$ predicted products \textit{separately for each country}, and then we average on the countries. This is of practical relevance from a policy perspective, because many new products appear in already highly diversified countries, while we would like to be precise also in low and medium income countries. By using mP@k we quantify the correctness of our possible recommendations of k products, on average, for a country.
\item \textbf{Recall} Recall is defined as the ratio between true positives and the sum of true positives and false negatives or, in other words, the total number of products that a country will export after $\Delta$ years \cite{powers2011evaluation}. So a high recall is associated with a low number of false negatives, that is, if we predict that a country will not start exporting a product, that country will usually not export that product. A negative recommendation is somehow less usual in strategic policy choices.
\item \textbf{F1 Score} The F1 score or F-measure \cite{dice1945measures,van1974foundation} is defined as the harmonic mean of precision and recall. As such, it is possible to obtain a high value of F1 only if both precision and recall are relatively high, so it is a very frequent choice to assess the general behavior of the classificator. As mentioned before, both precision and recall can be trivially varied by changing the scores' binarization threshold; however, the threshold that maximizes the F1 score is far from trivial, since precision and recall quantify different properties and are linked here in a non linear way. The \textbf{Best F1 Score} is computed by finding the threshold that maximizes the F1 score. 
\item \textbf{Area under the PR curve} It is possible to build a curve in the plane defined by precision and recall by varying the threshold that identifies the value above which the scores are associated to positive predictions. This value is not mislead by the class imbalance \cite{saito2015precision}.
\item \textbf{ROC-AUC} The Area Under the Receiving Operating Characteristic Curve \cite{hanley1982meaning,hajian2013receiver} is a widespread indicator that aims at measuring the \textit{overall} predictive power, in the sense that the user does not need to specify a threshold, like for Precision and Recall. On the contrary, all the scores are considered and ranked, and for each possible threshold both the True and the False Positive Rate (TPR and FPR, respectively) are computed. This procedure allows to define a curve in the TPR/FPR plane, and the area under this curve represents the probability that a randomly selected positive instance will receive an higher score than a randomly selected negative instance \cite{fawcett2006introduction}. For a random classifier, AUC=0.5 . It is well known \cite{saito2015precision,fernandez2018learning} that in the case of highly imbalanced data the AUC may give too optimistic results. This is essentially due to its focus on the overall ranking of the scores: in our case, misordering even a large number of not exported products does not affect the prediction performance; one makes correct true negative predictions only because there are a lot of negative predictions to make.
\item \textbf{Matthews coefficient} 
Matthews' correlation coefficient \cite{matthews1975comparison} takes in to account all the four classes of the confusion matrix and the class imbalance issue \cite{chicco2020advantages,boughorbel2017optimal}. 
\item \textbf{Accuracy} Accuracy is the ratio between correct predictions (true positives and true negatives) and the total number of predictions (true positives, false positives, false negatives and true negatives) \cite{powers2011evaluation}. In our prediction exercise we find relatively high values of accuracy essentially because of the overwhelming number of (trivially) true negatives (see table \ref{tab:algos}).
\item \textbf{Negative predictive value} Negative predictive value is defined as the ratio between true negatives and negatives, that are the products we predict will not be exported by a country \cite{powers2011evaluation}. Also in this case, a major role is played by the very large number of true negatives, that are however less significant from a policy perspective.
\end{itemize}
\subsection*{Libraries for the ML models}
Most of the models are implemented with scikit-learn 0.24.0 using the default parameters; in particular we used
\begin{itemize}
\item sklearn.ensemble.RandomForestClassifier
\item sklearn.svm.SVC
\item sklearn.linear\_model.LogisticRegression
\item sklearn.tree.DecisionTreeClassifier
\item sklearn.tree.ExtraTreeClassifier
\item sklearn.ensemble.AdaBoostClassifier
\item sklearn.naive\_bayes.GaussianNB
\end{itemize}
XGBoost is implemented using the library xgboost 1.2.0.\\
Finally, the neural network is implemented using keras 2.4.3. It consists on 2 layers with 32 neurons and activation function RELU and a final layer with a single neuron and sigmoid activation. We used rmsprop as optimizer, binary\_crossentropy as loss function and accuracy as loss metric.
\subsection*{Comparison with the work of O'Clery et al.}
In a recent paper, O’Clery et al. \cite{o2021productive} use export data from 1984 to 2009 to build a network of products called Eco Space and test the predictive power of the network by looking to the appearance of new products in the export of countries between 2010 and 2016. More specifically they consider the products that start with a RCA value less than 0.1 in 2010 and reach a value greater than 1 in the following years. Making this prediction, they obtain a ROC-AUC score of 0.715.\\
There are a number of differences between our data and theirs that make a direct comparison of their results with ours very hard. The main differences are i) they use a SITC 4 digit classification, while we use an HS 1992 6 digit classification and ii) they use a range of 26 years (from 1984 to 2009) to build the Eco Space, while we use only 18 years (from 1996 to 2013) to train our models.\\
To make a comparison between our models and the Eco Space predictive power we aggregated our export data to 4 digit and we trained a non cross validated random forest. We considered the possible activations of products with $RCA<0.1$ in 2013 and $RCA>1$ in 2018 obtaining a ROC value of 0.737.\\
Our score is higher, however we point out that, even after the 4 digit aggregation of our data, between the two datasets there are still significant differences and, as previously discussed, in the case of data with high class imbalance the ROC-AUC score is not the best choice to measure and compare prediction performances.
\bibliography{sample}

\end{document}